%% file: main.tex
\documentclass[sigconf]{acmart}

\usepackage{wrapfig}
\usepackage{amsmath}
\usepackage{booktabs}

\usepackage{graphicx}  
\usepackage{multirow}  
\usepackage{makecell}  
\usepackage{subcaption} 

\AtBeginDocument{%
  }

\setcopyright{acmlicensed}
\copyrightyear{2026}
\acmYear{2026}
\acmDOI{XXXXXXX.XXXXXXX}
\acmConference[ACM MM '26]{Proceedings of the 34th ACM International Conference on Multimedia}{November 10--14, 2026}{Rio de Janeiro, Brazil}
\acmISBN{978-1-4503-XXXX-X/2026/11}

\begin{document}

\title{PostCam: Camera-Controllable Novel-View Video Generation with Query-Shared Cross-Attention}

\author{Yipeng Chen}
\authornote{Equal Contribution.}
\affiliation{%
  \institution{State Key Lab of CAD\&CG, Zhejiang University}
  \country{China}}

\author{Zhichao Ye}
\authornotemark[1]
\affiliation{%
  \institution{Shanghai InSpatio Intelligent Technology Co., Ltd.}
  \country{China}}

\author{Zhenzhou Fan}
\affiliation{%
  \institution{State Key Lab of CAD\&CG, Zhejiang University}
  \country{China}}

\author{Xinyu Chen}
\affiliation{%
  \institution{State Key Lab of CAD\&CG, Zhejiang University}
  \country{China}}

\author{Xiaoyu Zhang}
\affiliation{%
  \institution{Shanghai InSpatio Intelligent Technology Co., Ltd.}
  \country{China}}

\author{Jialin Liu}
\affiliation{%
  \institution{Shanghai InSpatio Intelligent Technology Co., Ltd.}
  \country{China}}

\author{Nan Wang}
\affiliation{%
  \institution{Shanghai InSpatio Intelligent Technology Co., Ltd.}
  \country{China}}

\author{Guofeng Zhang}
\authornote{Corresponding Authors.}
\affiliation{%
  \institution{State Key Lab of CAD\&CG, Zhejiang University}
  \country{China}}

\author{Haomin Liu}
\affiliation{%
  \institution{Shanghai InSpatio Intelligent Technology Co., Ltd.}
  \country{China}}

\renewcommand{\shortauthors}{Chen et al.}

\begin{abstract}
We propose PostCam, a streamlined framework for novel-view video generation that achieves superior detail preservation and precise camera trajectory editing in dynamic scenes. Current methods often struggle with a trade-off between pose-based control, which lacks visual detail, and rendering-based guidance, which is overly sensitive to geometric accuracy. Despite recent hybrid attempts, achieving precise motion and visual consistency remains challenging due to the lack of effective cross-modal alignment.
We argue that robust control stems from the deep alignment of multimodal signals rather than increased input complexity. Our core contribution is the Query-Shared Cross-Attention mechanism, which projects 6-DoF poses and rendered features into a unified latent space. This allows the model to spontaneously achieve intrinsic consistency between motion cues and pixel-level guidance during denoising. Experiments demonstrate that PostCam maintains high-fidelity visual details while outperforming state-of-the-art methods by 20\% in trajectory precision, exhibiting superior robustness in complex dynamic scenes.
\end{abstract}

\keywords{novel-view video generation, camera control, video diffusion model, cross-attention}

\begin{teaserfigure}
  \includegraphics[width=\textwidth]{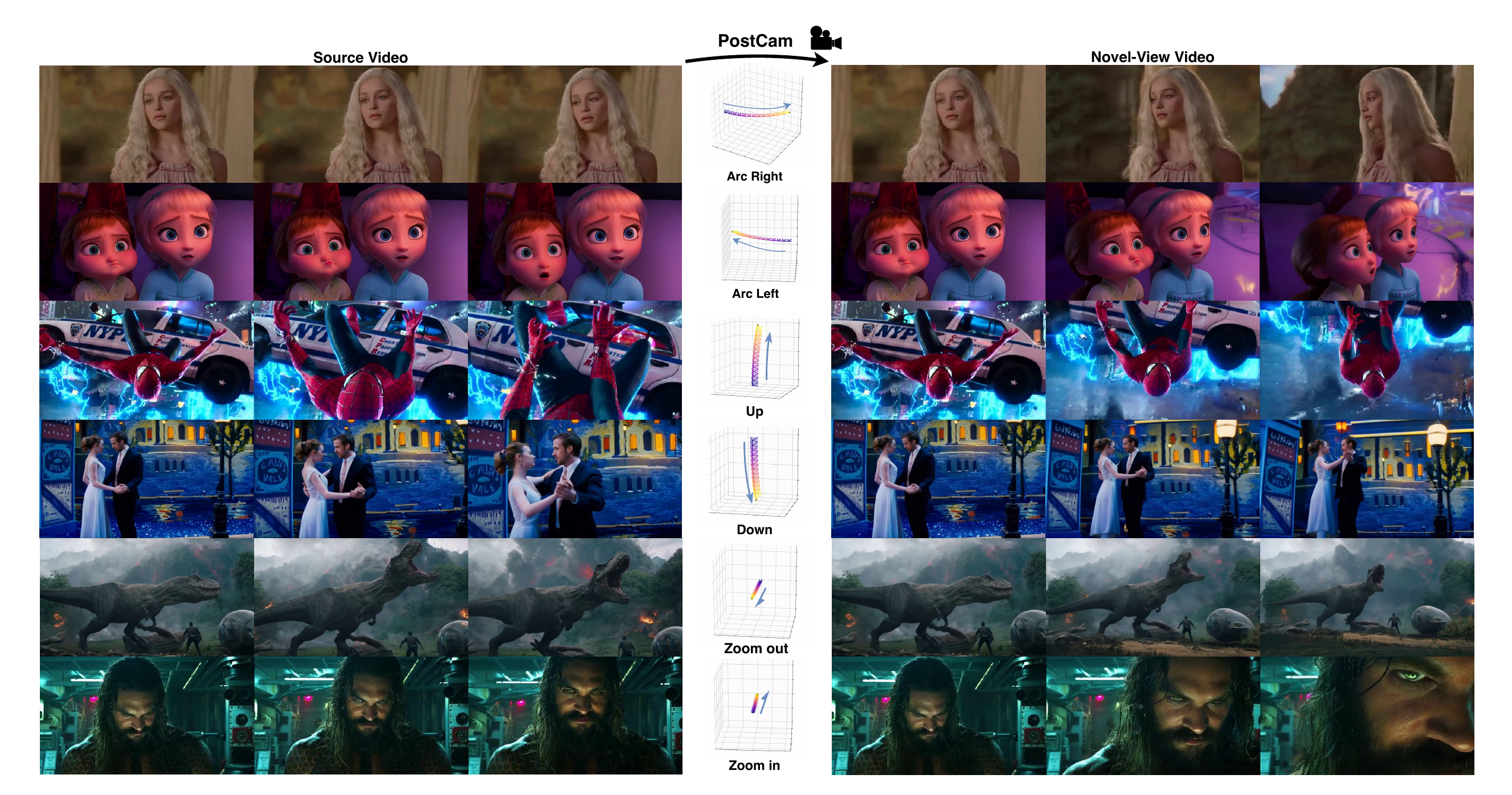}
  \caption{
Given a source video (left) and a target camera trajectory (middle), PostCam generates high-quality novel-view videos (right), enabling post-capture editing of camera motion in dynamic scenes.
Despite employing a lightweight backbone with only 1.3B parameters, PostCam achieves precise trajectory control and high-fidelity novel-view synthesis across a wide range of video styles and motion patterns.
  }
  \Description{Teaser figure showing PostCam results: source video, target camera trajectory, and generated novel-view videos across multiple scenes.}
  \label{fig:teaser}
\end{teaserfigure}

\maketitle

\input{paper/intro}

\input{paper/related_work}
\input{paper/method}

\input{paper/experiment}

\section{Conclusion}
We propose PostCam, a novel-view video generation framework for dynamic scenes that enables precise post-capture editing of camera trajectories. To address the limitations of existing methods, including low camera control accuracy and inadequate detail preservation, we introduce a query-shared cross-attention module combined with a two-stage training strategy, which significantly enhances the understanding and manipulation of camera motion.
Experiments on both real-world and synthetic datasets demonstrate that PostCam surpasses current state-of-the-art methods, achieving over 20\% improvement in camera control precision and view consistency while delivering the highest video generation quality.

\newpage

\bibliographystyle{ACM-Reference-Format}
\bibliography{main}
\newpage

\input{X_suppl}

\end{document}

%% file: paper/intro.tex
\begin{figure}[t]
  \centering
  \includegraphics[width=1\linewidth]{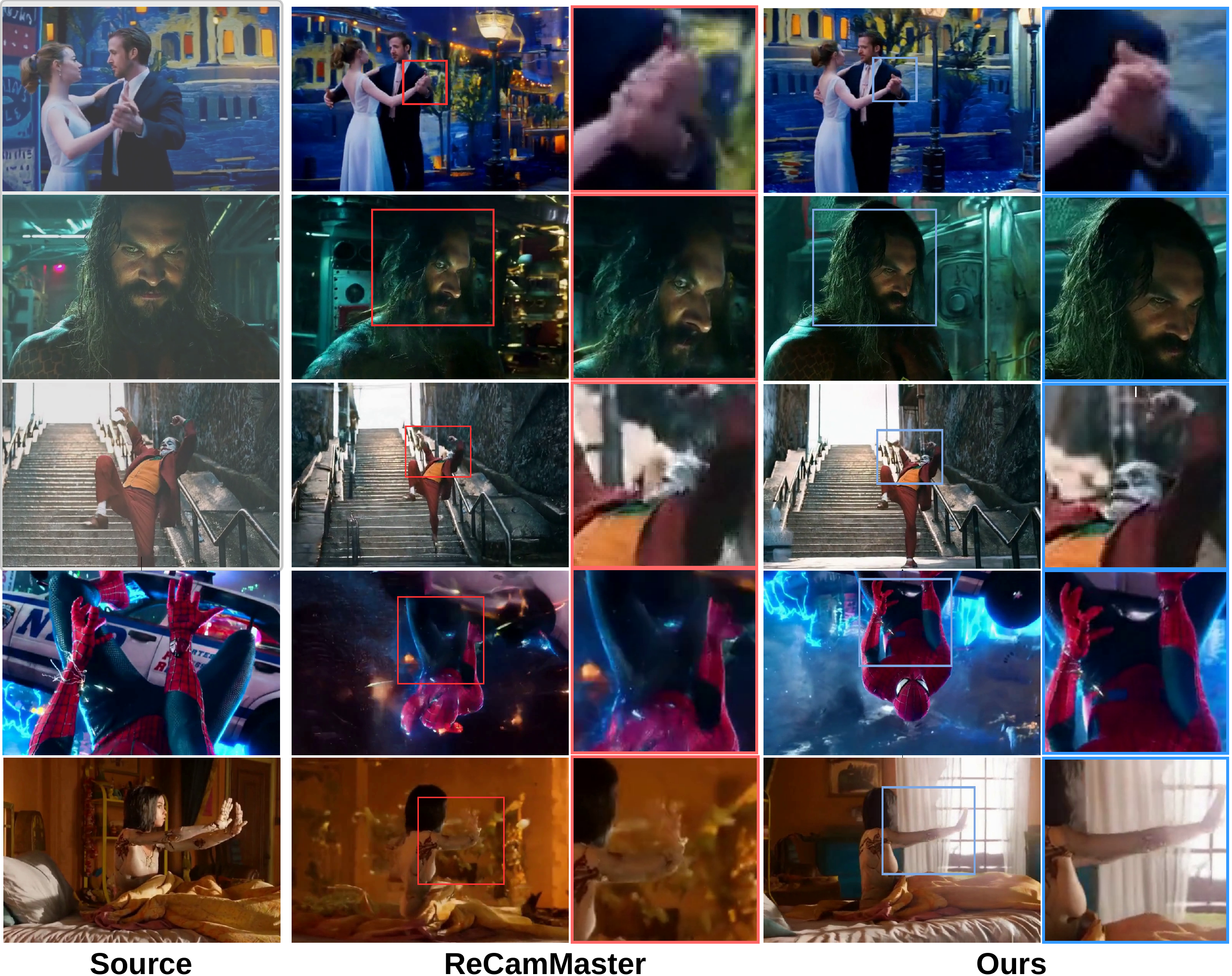}
  \caption{
\textbf{Comparison with State-of-the-Art Methods in Real Dynamic Scenes.} We compare the last-frame results of state-of-the-art methods with those of our approach, using the last frame of the source video (first column) as a reference.
Across all examples, our method consistently preserves fine details (e.g., faces, hands), whereas existing methods often suffer from blurring, distortion, and identity drift.
Moreover, our approach remains robust even in challenging cases where other methods fail completely (rows 4--5), producing high-quality and temporally coherent results.
For additional comparisons, please refer to our \textbf{supplementary videos}.
  }
  \label{fig:intro}
\end{figure}

\section{Introduction}

Recent years have witnessed remarkable progress in video generation models~\cite{peebles2023scalable,brooks2024video, yang2024cogvideox, wan2025wan, zheng2024open}, with substantial improvements in both fidelity and diversity. Consequently, research has increasingly focused on controllable video generation to meet the demands of professional production. Among various control signals, camera motion is fundamental as it defines the viewpoint trajectory and narrative rhythm. However, high-end camera effects typically require expensive hardware, making "post-capture editing" of camera trajectories a highly sought-after, yet challenging task that demands strict temporal synchronization and spatial consistency.

Current research in camera-controllable generation often oscillates between two technical paradigms, each with inherent limitations. Pose-based methods~\cite{bai2025recammaster} encode trajectories in numerical embeddings, which provides flexible control but often fails to preserve intricate visual details, leading to identity drift or blurriness. In contrast, rendering-based approaches~\cite{mark2025trajectorycrafter} lift source videos into 3D proxy representations for visual guidance; however, these methods are overly sensitive to the accuracy of geometric priors (e.g., depth estimation), where minor errors can trigger severe artifacts. Although recent hybrid attempts~\cite{cao2025mvgenmaster, cao2025uni3c} combine both signals via latent concatenation, they still suffer from significant motion deviations and detail inconsistencies due to the lack of an effective cross-modal alignment mechanism.

We argue that the key to achieving robust and precise control lies not in increasing the complexity of auxiliary inputs, but in the deep alignment of multimodal control signals. To this end, we propose PostCam, a streamlined and efficient framework for novel-view video generation. The core innovation is the Query-Shared Cross-Attention (QSCA) mechanism. By projecting 6-DoF poses and rendered features into a unified latent space, QSCA enables the model to spontaneously achieve intrinsic consistency between motion cues and pixel-level guidance during the denoising process. To facilitate better convergence, we further adopt a two-stage training strategy that decouples coarse motion learning from fine-grained appearance refinement.

Experimental results demonstrate that PostCam achieves superior detail preservation while outperforming state-of-the-art methods by over 20\% in trajectory precision. Our framework exhibits remarkable robustness in complex dynamic scenes, providing a high-fidelity solution for post-capture camera editing. 

In summary, our contributions are summarized as follows:
\begin{itemize}
\item We propose \textbf{PostCam}, a streamlined framework for novel-view video generation that enables precise camera trajectory editing while maintaining high-fidelity visual details in dynamic scenes.
\item We introduce the \textbf{Query-Shared Cross-Attention (QSCA)} mechanism, which achieves deep alignment between 6-DoF poses and rendered features within a unified latent space, ensuring intrinsic consistency in motion control.
\item We propose a \textbf{two-stage training strategy} that effectively decouples motion acquisition and appearance refinement, significantly enhancing  generation quality.
\item Extensive experiments show that PostCam outperforms state-of-the-art methods by 20\% in camera-control accuracy and view consistency, achieving superior performance on both real and synthetic datasets.
\end{itemize}

%% file: paper/related_work.tex
\section{Related Work}

\subsection{Video Generation.}
Our method builds upon video diffusion models, which have emerged as the prevailing paradigm for video generation. VDM~\citep{ho2022video} pioneered the application of diffusion models to high-resolution video synthesis.
In recent years, video diffusion architectures have transitioned from U-Net designs~\citep{blattmann2023stable, guo2023animatediff, singer2022make} to transformer-based architectures~\citep{brooks2024video, yang2024cogvideox, wan2025wan, kong2024hunyuanvideo, zheng2024open}, enabling greater realism and dynamic fidelity.
Among them, Wan2.1~\citep{wan2025wan} demonstrates superior generation capability as an open-source model and is therefore selected as our backbone.

\subsection{Camera-Controllable Video Generation.}
Camera-controllable video generation was initially advanced within the frameworks of text-to-video and image-to-video synthesis.
As a pioneering work, MotionCtrl~\citep{wang2024motionctrl} encodes 6-DoF camera poses and injects them into the diffusion model, fine-tuning on paired video-camera data to achieve video generation with arbitrary trajectory control.
Several subsequent works~\citep{he2024cameractrl, bahmani2024vd3d, kuang2024collaborative, zheng2024cami2v, liang2025wonderland, xu2024camco, bahmani2025ac3d, wang2024cpa} enhance camera accuracy by introducing Pl\"ucker embeddings to represent camera configurations.
Nevertheless, learning camera-motion control solely from numerical or Pl\"ucker representations is challenging, as it requires cross-modal interaction between visual information and numerical signals.
Beyond direct camera-pose conditioning, many approaches~\citep{yu2024viewcrafter, muller2024multidiff, ren2025gen3c, li2025realcam, feng2024i2vcontrol, popov2025camctrl3d, gu2025diffusion, zhai2025stargen} instead leverage point-cloud-rendered videos for camera-motion control: monocular depth is first lifted to a 3D point cloud, after which a proxy video is rendered from the target camera pose and injected into the network as a strong visual constraint.
DaS~\citep{gu2025diffusion} takes this further by replacing the RGB render with a 3D point-trajectory video, endowing the diffusion model with intrinsic 3D awareness. Some image-to-video methods, such as MVGenMaster~\cite{cao2025mvgenmaster} and Uni3C~\cite{cao2025uni3c}, attempt simple modal fusion by combining Pl\"ucker coordinates (from camera poses) with rendered videos in the latent space via addition and channel concatenation.
Furthermore, several training-free methods have been proposed~\citep{hou2024training, hu2024motionmaster, ling2024motionclone, xiao2024video}.
Overall, these methods achieve a degree of camera control, yet they remain restricted to reference-image or prompt-based generation and cannot synthesize results conditioned on a given source video.

\subsection{Novel-View Video Generation.}

Novel-view video generation with camera control has been pursued along two main avenues.
Pose-based methods condition a diffusion model on extrinsic camera parameters.
GCD~\cite{van2024generative} pioneers camera-controllable video-to-video generation by training on synthetically paired data from the Kubric simulator; however, its performance degrades under the large domain gap between synthetic and real videos.
ReCamMaster~\cite{bai2025recammaster} augments the noised latent with encoded pose vectors, yet the implicit injection inevitably perturbs visual quality.
A common limitation of these approaches is the weak cross-modal alignment between purely numerical pose signals and rich visual content, resulting in imprecise trajectories and limited fidelity.
Rendering-based methods~\citep{zhang2025recapture,xiao2024trajectory, bian2025gs, mark2025trajectorycrafter, yesiltepe2025dynamic} circumvent direct parameter injection by constructing an explicit 3D proxy.
TrajectoryCrafter~\cite{mark2025trajectorycrafter} lifts the source video into a point cloud via monocular depth estimation, renders a proxy video from the target camera, and uses this rendered video as a strong visual condition.
Although impressive control is achieved when the depth map is accurate, depth errors propagate into texture drift or ghosting artifacts that corrupt the final output.
These rendering-based methods~\citep{zhang2025recapture,xiao2024trajectory, bian2025gs, mark2025trajectorycrafter, yesiltepe2025dynamic} share the same fragility: imperfect or incomplete point clouds severely degrade video quality. In contrast to our task (which involves re-generating a video from a new viewpoint based on an input video), CameraCtrl II~\cite{he2025cameractrl} is dedicated to dynamic scene exploration, which it achieves by iteratively generating the next coherent video clip conditioned on the previous one. In summary, pose-based techniques suffer from weak visual--pose alignment, whereas rendering-based techniques are critically sensitive to depth accuracy. Both paradigms currently fail to simultaneously deliver robust pose control and high-fidelity results in real-world scenes.


%% file: paper/method.tex
\section{Method}

\begin{figure*}[t]
  \centering
  \includegraphics[width=0.8\textwidth]{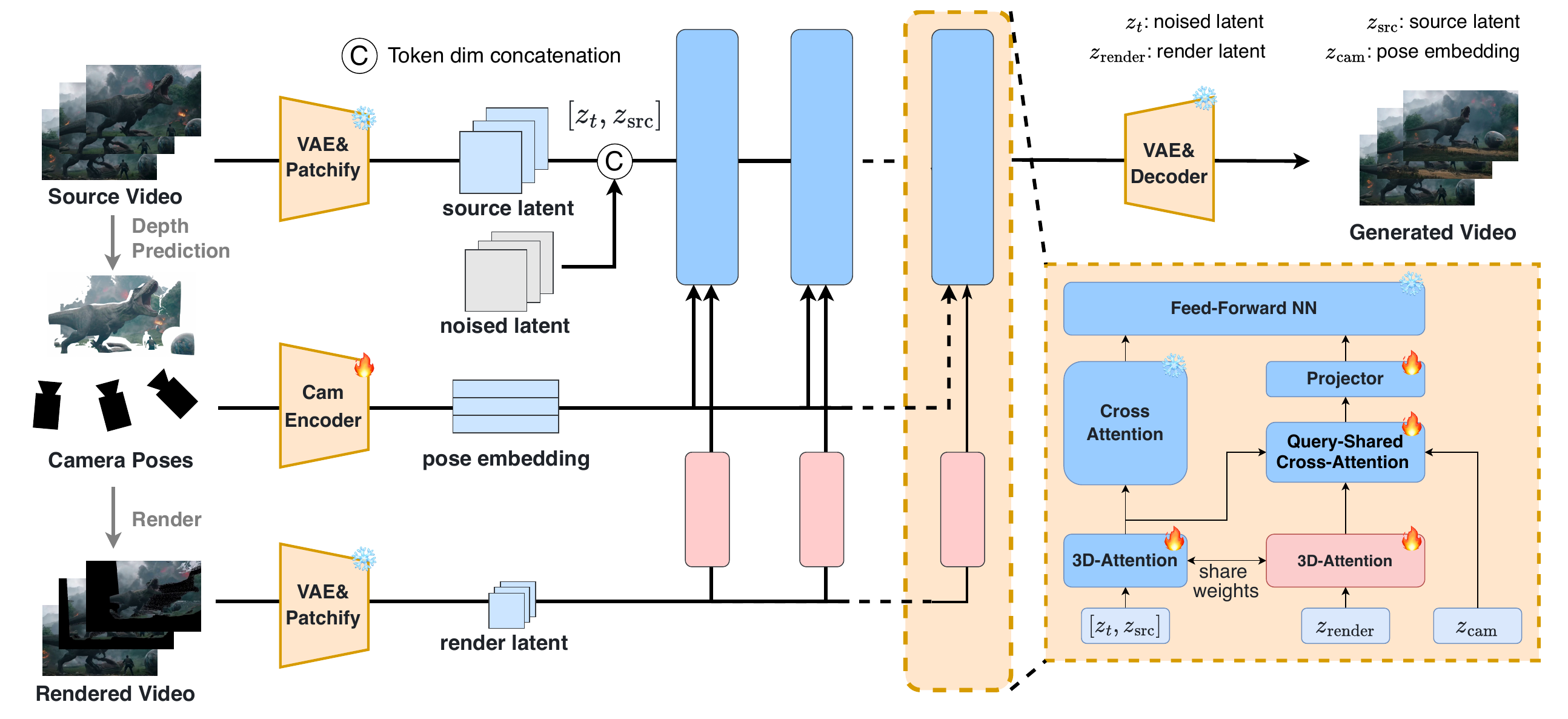}
  \caption{
\textbf{Framework overview}.
The model stacks multiple transformer blocks along three parallel pathways.
(a) The source video is first encoded into latent space and then concatenated with the noised latents before entering the transformer blocks.
(b) The camera parameters are processed by a lightweight encoder and injected into every block via query-shared cross-attention. The specific implementation details of this query-shared cross-attention mechanism are further elaborated in Fig.~\ref{fig:queryshared}.
(c) The rendered video is similarly encoded into latent space; within each block, it undergoes self-attention to extract high-level visual features.
An exploded view (bottom right) depicts the internal structure of a transformer block. 
  }
  \label{fig:framework}
\end{figure*}


\subsection{Preliminary: Base Model.}
We adopt Wan2.1~\citep{wan2025wan} as the backbone, a flow matching~\citep{lipman2022flow} framework built upon the prevailing diffusion transformer.
It comprises three key components: Wan-VAE, a diffusion transformer, and a text encoder.

Given an input video $x_0$, the Wan-VAE compresses it into the latent space $z_0$.
The model learns the conditional distribution $p(z_0|c_{\text{txt}})$, where $c_{\text{txt}}$ denotes the text prompt.
Wan2.1 employs flow matching to define the forward process with straight paths from the data distribution to the standard normal distribution:

\begin{equation}
    z_t = (1-t)z_0 + t\epsilon, \quad \epsilon \sim \mathcal{N}(0,I)
\end{equation}

where $t$ represents the diffusion timestep and the noised latent $z_t$ is constructed by adding random noise to $z_0$.
The model predicts the velocity field $\mathcal{V}_{\theta}$ through:

\begin{equation}
    \min_{\theta} \mathbb{E}_{z_0, t, \epsilon, c_{\text{txt}}} \left[ \left\| \mathcal{V}_{\theta}(z_t, t, c_{\text{txt}}) - (\epsilon - z_0) \right\|^2 \right]
\end{equation}

\subsection{Overview.}
The objective is to learn a model that generates a novel-view target video $x_{\text{tgt}}$, conditioned on a source video $x_{\text{src}}$ and a target camera trajectory $c_{\text{cam}}$.
Specifically, we model the conditional distribution $p(z_0\mid c_{\text{txt}}, c_{\text{cam}}, z_{\text{src}})$, where $z_0\in \mathbb{R}^{c\times (f\times h \times w)} $ is the latent representation of the target video, $z_{\text{src}}\in \mathbb{R}^{c\times (f\times h \times w)}$ is the latent of the source video, $c_{\text{txt}}$ denotes the text condition, and $c_{\text{cam}}$ represents the camera condition.
The symbols $c$, $f$, $h$, and $w$ denote the channel count, temporal frames, spatial height, and width of the latent features after VAE encoding and patch tokenization.

As shown in Fig.~\ref{fig:framework}, the text-to-video backbone is extended with three additional inputs: the source video, the camera trajectory, and a rendered video.
The source video, serving as a visual reference for both appearance and motion dynamics, is first encoded by a variational autoencoder and tokenized into patches. The resulting latent representation is concatenated with the noised latent along the token dimension to form $[z_t, z_{\text{src}}] \in \mathbb{R}^{c \times (2f \times h \times w)}$,
and the concatenated latent is processed by a self-attention mechanism to extract high-level visual representations.
The camera conditions consist of two modalities: the rendered video and the camera trajectory.
The proposed query-shared cross-attention jointly injects both modalities into the noised latent, effectively incorporating camera motion information during denoising.

The training objective is given by:
\begin{align}
\min_{\theta}\, \mathbb{E}_{z_0, t, \epsilon, c_{\text{txt}}, c_{\text{cam}}, z_{\text{src}}} \Bigg[
    &\left\| \mathcal{V}_{\theta}(z_t,\, t,\, c_{\text{txt}},\, c_{\text{cam}},\, z_{\text{src}}) \right. \notag\\
    &\left. {} - (\epsilon - z_0) \right\|^2
\Bigg]
\end{align}

\subsection{Camera Condition Preprocessing.}

Before injecting the camera conditions into the noised latents, we perform distinct preprocessing for the camera parameters and the rendered video.

For the camera poses, a lightweight camera encoder is trained to map the camera extrinsic matrix $\mathbf{c}_{\text{cam}}$ to a pose embedding $\mathbf{z}_{\text{cam}}\in\mathbb{R}^{c\times f}$.
To preserve temporal consistency, 1D Rotary Positional Embeddings (RoPE) are applied during the subsequent cross-attention stage.

The rendered video is generated by first lifting the source video into 3D via monocular depth estimation and then reprojecting it along the target camera trajectory.
While existing methods rely on full-resolution rendering, we observe that within our query-shared mechanism, the primary role of the rendered video is to refine motion accuracy and enhance visual fidelity, rather than to achieve precise pixel-wise correspondence. This also explains why the model can still produce correct motion even when inaccurate depth maps lead to severely distorted rendering. Therefore, we downsample the rendered video in the spatial dimensions, reducing computational cost without compromising the quality of the generated output.
The downsampled rendered video is encoded together with the source video by the same VAE, after which high-level visual features are extracted through 3D self-attention, yielding $z_{\text{render}} \in \mathbb{R}^{c\times (f\times \frac{h}{4} \times \frac{w}{4})}$.
This self-attention module reuses the weights of the backbone without introducing additional parameters.

\begin{figure}[t]
    \centering
    \includegraphics[width=0.3\textwidth]{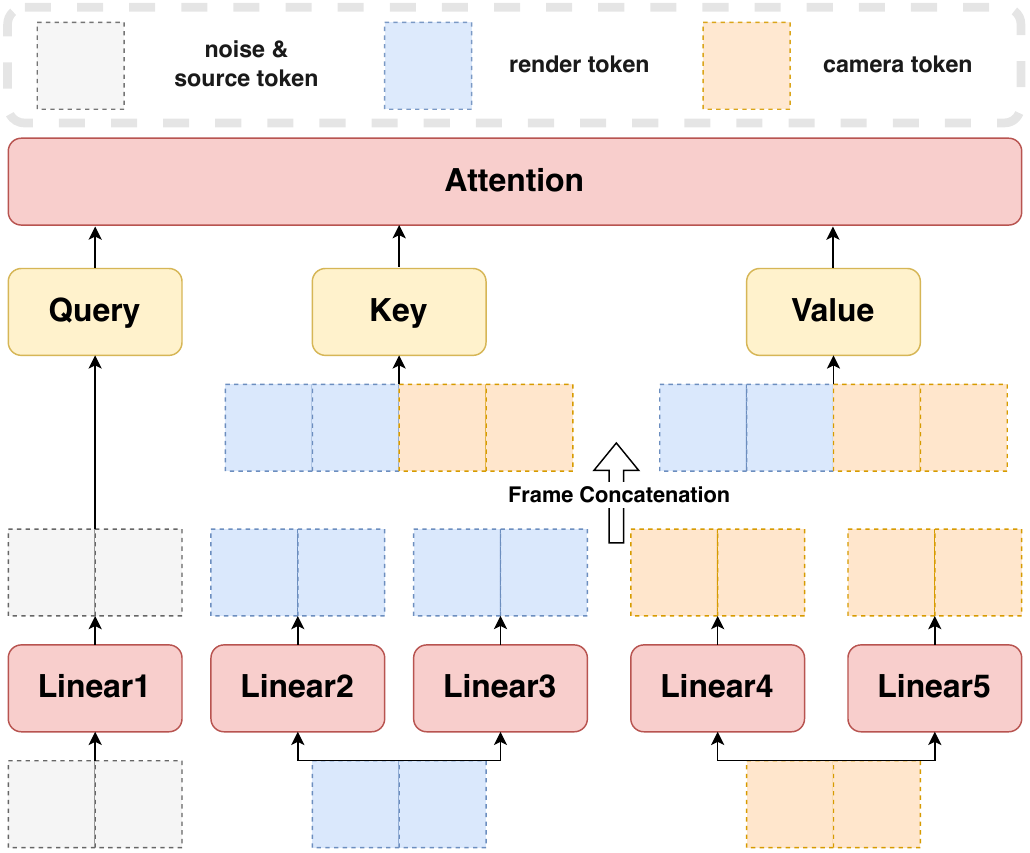}
    \caption{Query-shared cross-attention mechanism.}
    \label{fig:queryshared}
\end{figure}

\subsection{Query-Shared Cross-Attention}

To suppress extraneous cues from the dual-modality camera condition, we propose a novel query-shared cross-attention module.
Within each transformer block, the camera condition is injected into the noised latent via the proposed cross-attention mechanism.

In query-shared cross-attention, the query vector is obtained by applying a linear projection to $z_t$:
\[
Q = \mathrm{Linear}(z_t).
\]
Simultaneously, $z_{\text{render}}$ and $z_{\text{cam}}$ are each linearly projected to produce the corresponding keys ($ K_{\text{render}}$ and $K_{\text{cam}} $) and values ($ V_{\text{render}}$ and $V_{\text{cam}} $).

The keys and values from both conditions are concatenated along the token dimension, and attention is computed as:
\begin{align}
A &= \text{softmax}\left(
        \frac{Q\ [K_{\text{render}},\ K_{\text{cam}}]^{\top}_{\text{token-dim}}}{\sqrt{d_q}}
    \right)
\end{align}
where $d_q$ is the dimensionality of the query.
The module first computes an attention matrix from the query and the concatenated key features. A subsequent softmax operation produces a weighted distribution, assigning higher importance to features most relevant to the video generation task. These weights govern the selective aggregation of information from the concatenated value features, allowing the module to distill common camera motion cues for both precise pose control and high-fidelity generation.
The resulting attention output is projected and added back to the noised latent:
\begin{align}
z_t &= z_t + \text{projector}\Big(
    A\ [V_{\text{render}},\ V_{\text{cam}}]_{\text{token-dim}}
\Big)
\end{align}


\subsection{Training Strategy}
To preserve the video-generation capability of the base model, almost all parameters are frozen. Only the self-attention modules are updated to integrate information from the source latent into the noised latent, ensuring spatiotemporal consistency between them.
Additionally, we introduce three trainable modules beyond the base architecture: a camera encoder, a query-shared cross-attention block, and a lightweight projector.

In practice, we adopt a two-stage training strategy that enables the model to learn low-frequency camera-motion cues while avoiding interference from rendered visual signals during the early training phase:
\textit{Stage 1: Pose-Only Conditioning Phase.}
In this phase, only the camera extrinsic parameters are used; no rendered video is provided.
The camera features are injected as:
\begin{equation}
z_t = z_t + \text{projector}\left( \text{softmax}\left(\frac{Q K_{\text{cam}}^{T}}{\sqrt{d_q}}\right) V_{\text{cam}} \right)
\end{equation}
The projection layer is initialized to zero, effectively suppressing the influence of the camera-control signal during early training.
This allows the model to first learn content injection from the source video before acquiring camera-motion control capability. Without the projector module, the model struggles to converge, as it must learn both source injection and camera motion extraction from the extrinsic parameters simultaneously. \textit{Stage 2: Joint Conditioning Phase.}
In this phase, after the model has acquired coarse camera-control ability, both camera poses and rendered video are provided during training.
The model now benefits from the complementary nature of the two conditions, leading to improved control precision and overall generation quality.

%% file: paper/experiment.tex
\input{table/compare_sota_openvid}

\input{table/compare_sota_ourblender}

\section{Experimental Results}

\subsection{Experimental Setup.}
\noindent\textbf{Training Details.}
For a fair comparison, the model is trained on the same synthetic dataset as ReCamMaster~\cite{bai2025recammaster}, with a learning rate of $1 \times 10^{-5}$. The training is conducted on 8 NVIDIA H200 GPUs with a batch size of 2 per GPU, resulting in a total batch size of 16. The training process consists of two phases: in the first phase, the model is trained exclusively with camera poses until loss convergence; in the second phase, it undergoes joint training with both camera poses and rendered videos until reaching a final converged state. The additional parameters introduced include a linear projector and a cross-attention module within each DiT block, alongside a lightweight camera encoder composed of four linear layers. Following the experimental setup of ReCamMaster, we employ Wan2.1-1.3B~\citep{wan2025wan} as our base model. Since most state-of-the-art video generation backbones share similar DiT-style architectures, our method is inherently generalizable to other foundational frameworks.

\noindent\textbf{Evaluation Dataset.}
To enable accurate and comprehensive evaluation, we construct both real and synthetic test sets. The real-world set includes 100 clips from OpenViD~\citep{nan2024openvid}, each paired with 10 basic camera trajectories following the testing protocol of ReCamMaster, resulting in 1{,}000 test samples.
Since ground-truth videos are unavailable in real-world scenes, we additionally generate a synthetic test set following the pipeline of ReCamMaster, enabling direct comparison between generated and ground-truth videos.
Consisting of 100 samples across ten dynamic scenes, the synthetic dataset features intricate composite trajectories, in contrast to the ten basic camera movements. Examples and further details can be found in the supplementary material.

\noindent\textbf{Evaluation Metrics.}
\textit{(1) Camera Accuracy:}
To evaluate the camera motion control accuracy of different models, we estimate the camera trajectory of the generated video and compare it with the ground-truth trajectory, computing the rotation error (RotErr) and translation error (TransErr).
Since conventional pose estimation methods (e.g., COLMAP and GLOMAP) often fail in dynamic scenes, we evaluated several alternative approaches and ultimately adopted VGGT, a feed-forward reconstruction method that proved more suitable for our experiments. Detailed comparative results of different trajectory estimation methods are provided in the supplementary material.
\textit{(2) Video Quality:}
We evaluate video quality using the widely adopted VBench~\citep{huang2024vbench} metrics, reporting metrics across multiple dimensions, including Aesthetic Quality (Aes. Qual.), Imaging Quality (Img. Qual.), Temporal Flickering (Tem. Flick.), Motion Smoothness (Motion Smooth.), Subject Consistency (Sub. Cons.), and Background Consistency (Bg. Cons.).
\textit{(3) View Consistency:}
We assess view consistency between generated and real videos in the feature space using Fr\'echet Image Distance (FID)~\citep{heusel2017gans} and Fr\'echet Video Distance (FVD)~\citep{unterthiner2018fvd}.
For real-world datasets, we measure the similarity between generated videos and their corresponding source videos.
For synthetic datasets, we compare directly against ground-truth target videos.

\begin{figure*}[ht]
\centering
  \includegraphics[width=0.9\linewidth]{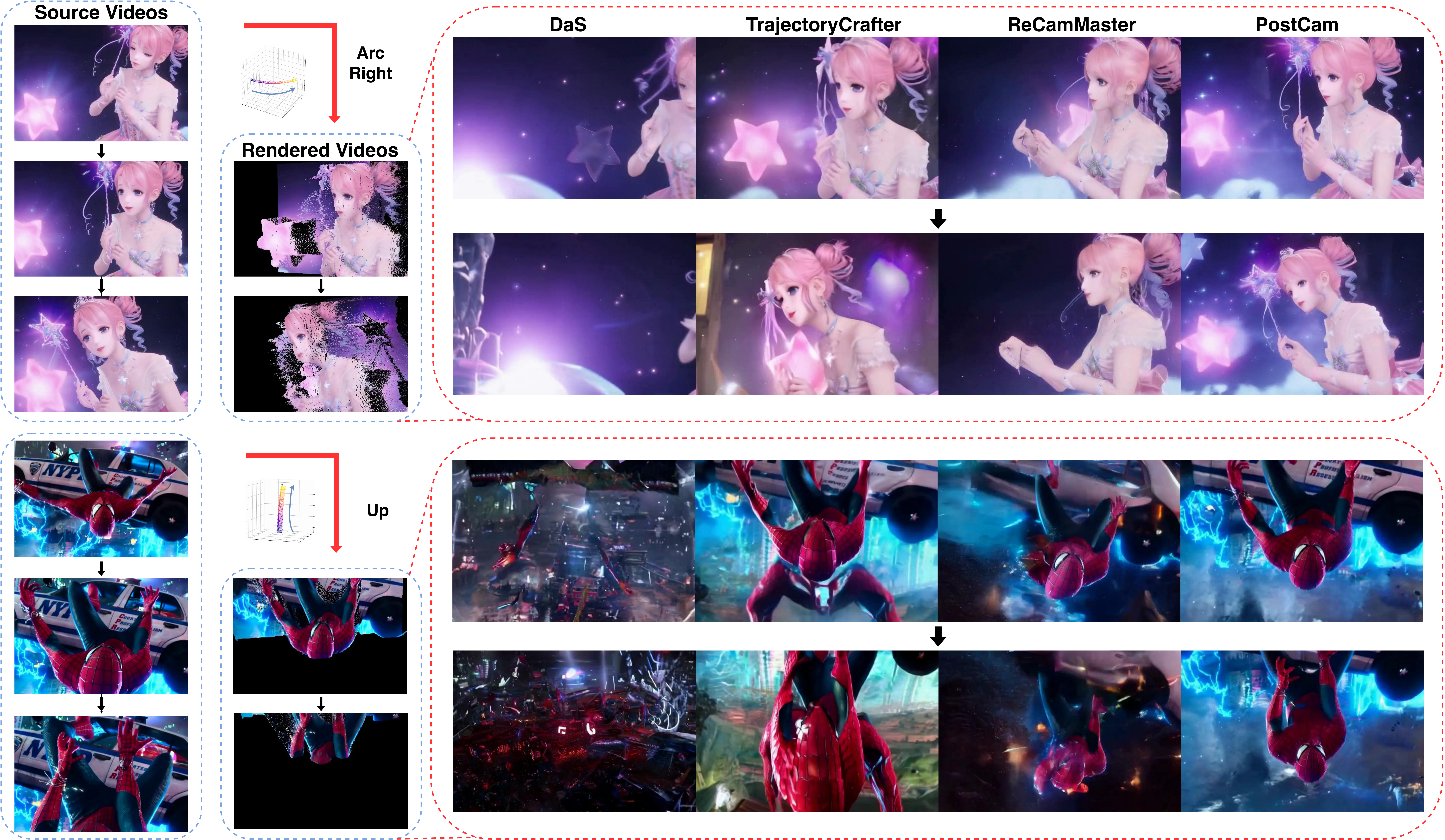}
  \caption{\textbf{Qualitative comparison results.}
Column 1 displays source frames (initial, middle, final), while column 2 illustrates target camera motions (direction and rendering). Columns 3--6 compare our method against leading approaches. As an I2V model, DaS fails to maintain temporal coherence. TrajectoryCrafter is sensitive to rendering quality, showing artifacts when depth data is inaccurate or incomplete (e.g., the missing head in Scene 2). While ReCamMaster is render-agnostic, it fails to balance visual quality with motion precision in complex scenarios. In contrast, our method robustly generates high-quality results that faithfully align with source content and achieve superior pose accuracy. 
}
  \label{fig:sota_compare_film}
\end{figure*}

\begin{figure}[t]
  \centering
  \includegraphics[width=\linewidth]{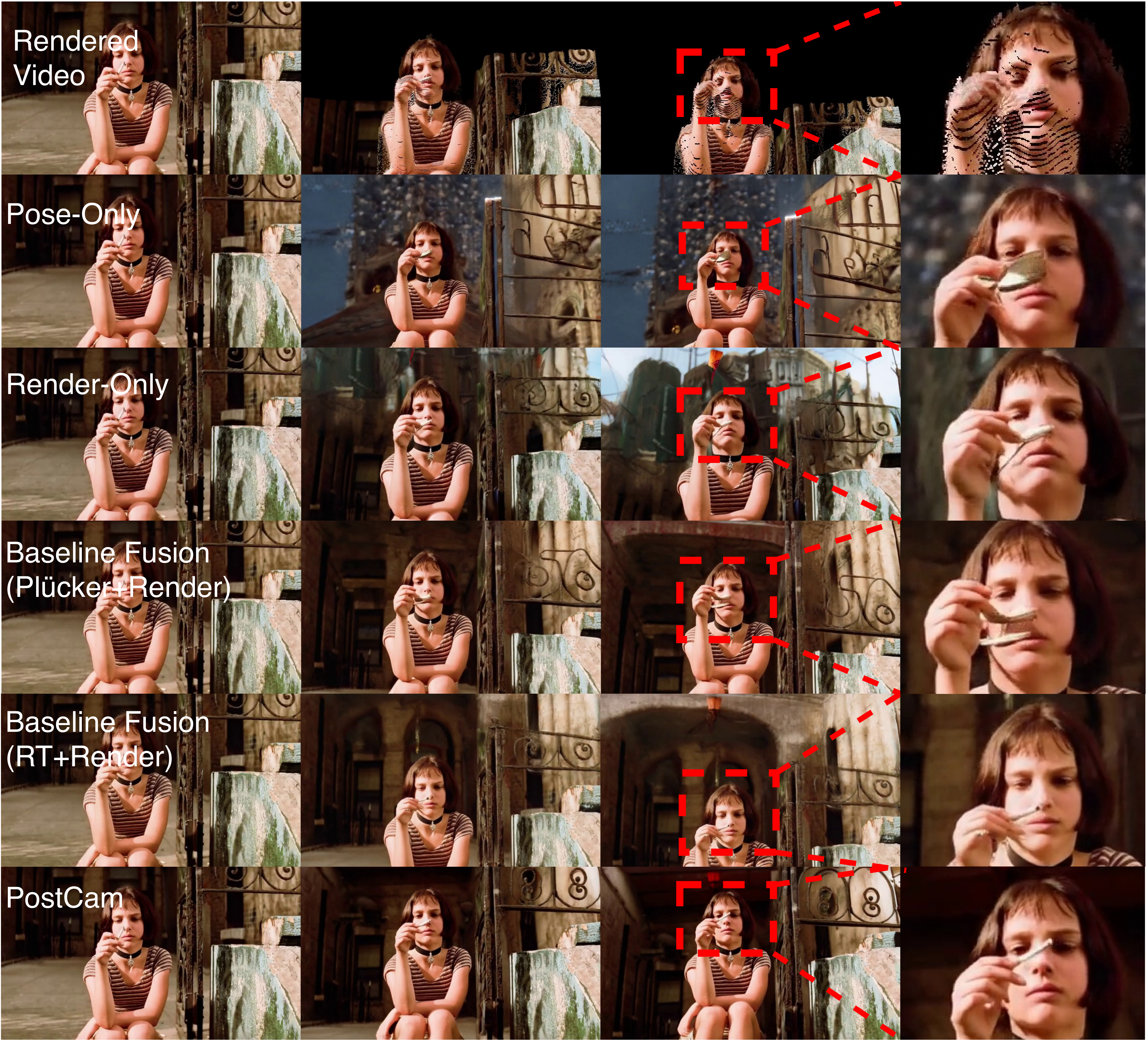}
  \caption{
  \textbf{Comparisons under different conditions}.
Our method achieves faithful preservation of fine details and more accurate background generation while ensuring accurate camera motion control. The red boxes highlight that, even when distortions occur in the rendered views, only our results retain fine details, such as the cigarette in the hand, while achieving superior pose accuracy.
  }
  \vspace{-3mm}
\label{fig:posevsrender}
\end{figure}

\subsection{Comparison with State-of-the-Art Methods}

We compare PostCam with state-of-the-art camera-conditioned video generation methods: DaS~\cite{gu2025diffusion}, TrajectoryCrafter~\cite{mark2025trajectorycrafter}, and ReCamMaster~\cite{bai2025recammaster}.
To ensure a fair evaluation, a unified preprocessing pipeline is applied to all generated videos to standardize their spatial resolution and temporal length. Specifically, all baselines are first generated using their official configurations (e.g., ReCamMaster at 81 frames) and subsequently normalized to a uniform duration of 41 frames through frame sampling. For spatial consistency, we crop and resize all outputs to a $480 \times 720$ resolution, following the protocol of DaS due to its minimal aspect ratio. This ensures that the initial frames across all methods contain identical visual information, preventing extraneous content from biasing the comparison.

As shown in Table~\ref{tab:compare_sota_openvid} and Table~\ref{tab:compare_sota_ourblender}, PostCam substantially outperforms the baseline methods in camera-control accuracy while simultaneously achieving state-of-the-art performance on most video-quality metrics.
Qualitative results in Fig.~\ref{fig:sota_compare_film} further highlight these advantages. Additional qualitative comparisons, including examples on OpenViD, synthetic data, and in-the-wild scenes, can be found in the supplementary materials and supplementary video.

\subsubsection{Robustness to Depth Noise}
\label{robustness}
\begin{figure}[h]
    \centering
    \includegraphics[width=1.0\linewidth]{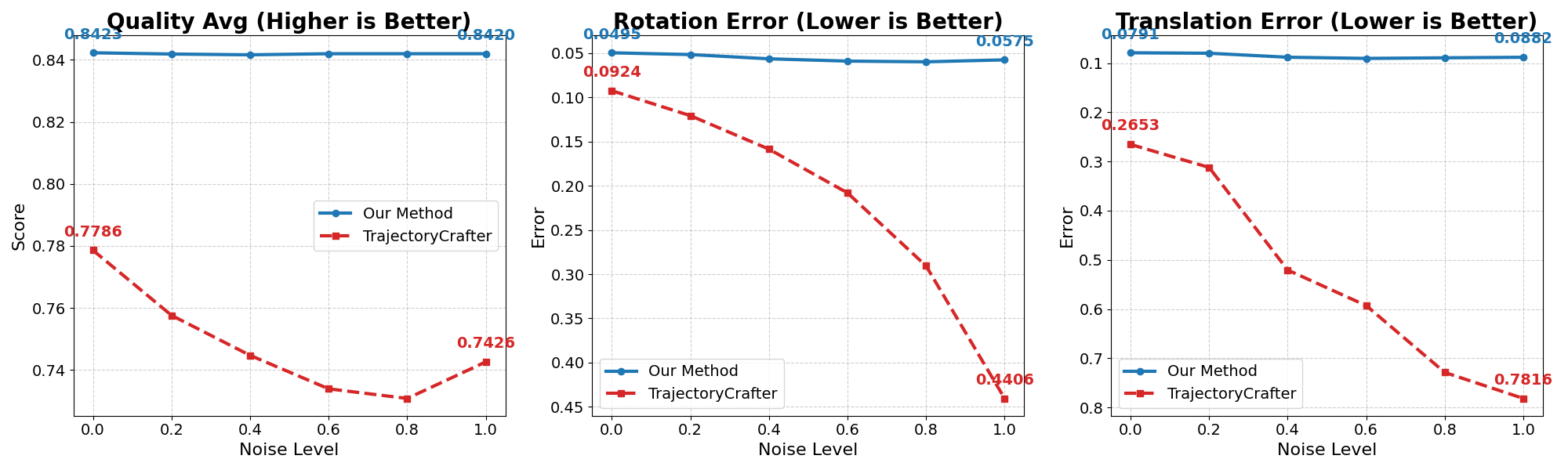}
    \caption{Comparison of VBench quality average (left), rotation error (middle), and translation error (right) under varying depth noise levels.}
    \label{fig:noise_robustness}
    \vspace{-3mm}
\end{figure}

To evaluate the robustness to depth noise, we add Gaussian noise of varying strengths to the point cloud before rendering.
As shown in Fig.~\ref{fig:noise_robustness}, our method remains highly robust, whereas the performance of TrajectoryCrafter degrades significantly.

\begin{figure}[h]
    \centering
    \includegraphics[width=1.0\linewidth]{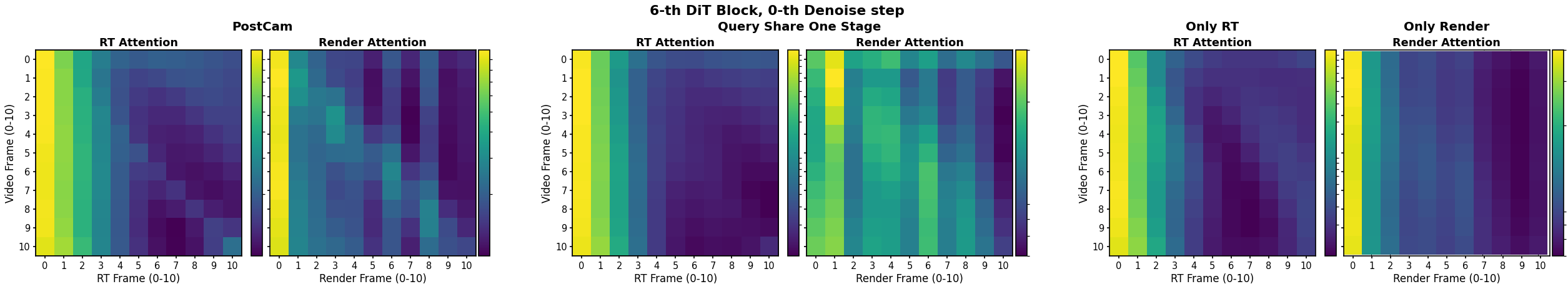}
    \caption{Attention Distribution Comparison.}
    \label{fig:attn_comparison}
    \vspace{-3mm}
\end{figure}

\subsubsection{User Study} We conduct a blind user study based on Mean Opinion Score (MOS) with 20 participants, comparing PostCam against ReCamMaster, TrajectoryCrafter, and DaS. Participants rate the videos on a scale of 1--5 across three dimensions: visual quality, camera controllability, and temporal consistency. As shown in Table~\ref{tab:user_study}, PostCam outperforms all baselines across all dimensions, validating that the improvements in automated metrics are consistent with human perception.



\subsection{Ablation Study}

\subsubsection{Ablation on Conditioning Strategies}

To evaluate the impact of different conditioning strategies, we implement the following variants under identical training settings:

\textit{Pose-only}: uses only camera parameters as the control signal, injected via cross-attention.

\textit{Render-only}: uses only the rendered video, encoded with 3D self-attention and injected via cross-attention.

\textit{Baseline Fusion (RT+Render)}: employs both camera parameters and rendered video, where each modality is encoded independently and injected into the network through its own dedicated cross-attention layer in a single stage.

\textit{Baseline Fusion (Pl\"ucker+Render)}: uses Pl\"ucker coordinates and rendered video. Following several image-to-video methods~\cite{cao2025mvgenmaster, cao2025uni3c} that fuse camera poses and rendered frames, we implement a video-to-video version in which the Pl\"ucker information is added to the noised latents and the rendered video is injected through cross-attention in a single stage.

Our main model is \textit{PostCam (RT+Render)}. We additionally implement a variant, \textit{PostCam (Pl\"ucker+Render)}, which uses Pl\"ucker coordinates as input. This variant also adopts the query-shared cross-attention module and the two-stage training strategy.

Table~\ref{tab:ablation_render_rt} demonstrates that a direct combination of both signals is suboptimal. Notably, the RT+Render baseline fails to converge effectively, resulting in the least accurate motion control among all tested configurations under equal training iterations.
We attribute this to the significant disparity between pose trajectories and rendered video information. Simultaneously injecting these two heterogeneous signals makes it difficult for the model to converge.
Converting camera poses into Pl\"ucker coordinates for injection partially alleviates the convergence issue---this also explains why most I2V methods adopt Pl\"ucker coordinates. However, this approach remains a suboptimal injection strategy.

\input{table/ablation_RT_Render_openvid}

In contrast, our method, which employs query-shared cross-attention, effectively extracts the essential and shared camera motion information from different modalities. This mechanism is agnostic to the pose representation, functioning robustly regardless of whether poses are expressed as camera parameters (RT) or Pl\"ucker coordinates. As shown in the table, our approach achieves state-of-the-art performance in both camera control accuracy and generation quality.

Specifically, through a single shared query, the model selectively extracts the most beneficial components from both modalities: it first learns fundamental motion patterns solely from camera poses and then incorporates richer visual cues from rendered videos for fine-grained control and quality enhancement. This progressive learning strategy, combined with the shared query mechanism, effectively resolves the aforementioned convergence difficulties and modal conflicts. The qualitative results in Fig.~\ref{fig:posevsrender} further corroborate our findings.

\subsubsection{Ablation on Query-Shared Cross-Attention}

To quantify the contribution of the proposed query-shared cross-attention, we ablate three condition injection strategies while keeping the training settings (including the two-stage training strategy) and input data identical.
\textit{w/o query-shared}: each transformer block employs two separate cross-attention layers, one for camera parameters and another for the rendered condition.
\textit{w/o kv-concatenated}: the query matrix is shared, but the keys and values of the two conditions are processed independently and injected into the noise via the same query.
\textit{Ours}: the query matrix is shared and the key--value pairs of both conditions are concatenated before attention computation.
For a fair comparison, all strategies undergo the same two-stage training pipeline.
As reported in Table~\ref{tab:ablation_qs}, the proposed query-shared cross-attention achieves the highest camera-motion control accuracy.
Under the shared-query design, the module first computes an attention matrix from the query and the concatenated key features. A subsequent softmax operation produces a weighted distribution that assigns higher importance to features most relevant to the video generation task. These weights govern the selective aggregation of information from the concatenated value features, enabling the module to distill common camera motion cues for both precise pose control and high-fidelity generation.
Qualitative results can be found in the supplementary material.

\input{table/ablation_div_qkv_openvid}

\input{table/ablation_train_strategy}


\subsubsection{Ablation on Two-Stage Training.}

To validate the efficacy of the proposed two-stage training, we ablate three strategies: (1) \textit{one stage}, which directly trains on both camera-control conditions; (2) \textit{two stage, render-first}, which employs only the rendered condition in the first stage and both conditions in the second; and (3) \textit{two stage, pose-first} (i.e., ours), which employs only the pose condition in the first stage and both conditions in the second.
As shown in Table~\ref{tab:ablation_train}, our two-stage training achieves the best camera control among the three variants.
In the one-stage case, the model must simultaneously learn from multiple control conditions, leading to suboptimal camera-control accuracy.
In the render-first two-stage case, the model tends to enforce pixel-level alignment between the rendered frames and the synthesized output rather than learning the underlying camera motion, resulting in degraded pose controllability.
By contrast, our approach enables the model to first learn the underlying camera motion from poses alone and subsequently incorporate rendered videos for finer motion control and enhanced visual fidelity.

\subsubsection{Attention Distribution Comparison}


The attention maps in Fig.~\ref{fig:attn_comparison} validate the contributions of our full PostCam model. PostCam (left) exhibits the clearest diagonal structure, signifying precise and temporally consistent cross-frame attention. This structured pattern progressively breaks down as key components are removed. Eliminating the two-stage training (Query Share One Stage, middle) noticeably degrades the diagonal. Furthermore, when both two-stage training and query sharing are removed (Only RT and Only Render, right), the diagonal structure is nearly eradicated, resulting in highly diffuse attention maps. This lack of focused temporal structure directly visualizes how these ablated models are significantly more prone to overfitting to irrelevant signals.



%% file: table/compare_sota_openvid.tex
\begin{table*}[ht]
\centering
\renewcommand{\arraystretch}{1.2}

\resizebox{\textwidth}{!}{
\begin{tabular}{l|cc|cccccc|cc}
\hline
\multirow{3}{*}{\textbf{Method}}
& \multicolumn{2}{c|}{\textbf{Camera Accuracy}}
& \multicolumn{6}{c|}{\textbf{Video Quality}}
& \multicolumn{2}{c}{\textbf{View Consistency}} \\
\cline{2-3} \cline{4-9} \cline{10-11}
    & \makecell[c]{Rot\\Err$\downarrow$}
    & \makecell[c]{Trans\\Err$\downarrow$}
    & \makecell[c]{Aes.\\Qual.$\uparrow$}
    & \makecell[c]{Img.\\Qual.$\uparrow$}
    & \makecell[c]{Tem.\\Flick.$\uparrow$}
    & \makecell[c]{Motion\\Smooth.$\uparrow$}
    & \makecell[c]{Sub.\\Cons.$\uparrow$}
    & \makecell[c]{Bg.\\Cons.$\uparrow$}
    & \makecell[c]{FID$\downarrow$}
    & \makecell[c]{FVD$\downarrow$}\\
\hline
DaS~\citep{gu2025diffusion}&
0.0724&
0.4872&
52.43&
61.69&
\textbf{96.98}&
98.94&
93.39&
92.56&
101.75&
474.29\\
TrajectoryCrafter~\citep{mark2025trajectorycrafter}&
0.0851&
0.2154&
51.69&
59.53&
95.31&
98.15&
90.37&
91.28&
114.71&
478.55\\
ReCamMaster~\citep{bai2025recammaster}&
0.0649&
0.1480&
52.23&
62.61&
96.15&
98.89&
92.97&
91.89&
85.68&
421.97\\
\textbf{PostCam}&
\textbf{0.0501}&
\textbf{0.1021}&
\textbf{53.52}&
\textbf{65.71}&
96.08&
\textbf{99.01}&
\textbf{94.03}&
\textbf{92.58}&
\textbf{67.20}&
\textbf{380.96}\\
\hline
\end{tabular}}
\caption{Quantitative comparison with state-of-the-art methods on the OpenViD dataset.}
\label{tab:compare_sota_openvid}
\end{table*}

%% file: table/compare_sota_ourblender.tex
\begin{table*}[ht]
\centering
\renewcommand{\arraystretch}{1.2}
\resizebox{\textwidth}{!}{
\begin{tabular}{l|cc|cccccc|cc}
\hline
\multirow{3}{*}{\textbf{Method}}
& \multicolumn{2}{c|}{\textbf{Camera Accuracy}}
& \multicolumn{6}{c|}{\textbf{Video Quality}}
& \multicolumn{2}{c}{\textbf{View Consistency}} \\
\cline{2-3} \cline{4-9} \cline{10-11}
    & \makecell[c]{Rot\\Err$\downarrow$}
    & \makecell[c]{Trans\\Err$\downarrow$}
    & \makecell[c]{Aes.\\Qual.$\uparrow$}
    & \makecell[c]{Img.\\Qual.$\uparrow$}
    & \makecell[c]{Tem.\\Flick.$\uparrow$}
    & \makecell[c]{Motion\\Smooth.$\uparrow$}
    & \makecell[c]{Sub.\\Cons.$\uparrow$}
    & \makecell[c]{Bg.\\Cons.$\uparrow$}
    & \makecell[c]{FID$\downarrow$}
    & \makecell[c]{FVD$\downarrow$}\\
\hline
DaS~\citep{gu2025diffusion}&
0.1112&
0.9010&
49.41&
52.62&
\textbf{96.25}&
98.67&
84.55&
89.58&
363.52&
988.56\\
TrajectoryCrafter~\citep{mark2025trajectorycrafter}&
0.0924&
0.2653&
51.64&
50.77&
94.11&
97.62&
83.54&
89.50&
250.91&
799.15\\
ReCamMaster~\citep{bai2025recammaster}&
0.0638&
0.1437&
58.50&
63.31&
95.98&
99.08&
91.13&
91.65&
103.74&
434.34\\
\textbf{PostCam}&
\textbf{0.0495}&
\textbf{0.0791}&
\textbf{59.80}&
\textbf{64.64}&
96.17&
\textbf{99.13}&
\textbf{92.48}&
\textbf{93.15}&
\textbf{80.29}&
\textbf{313.02}\\
\hline
\end{tabular}}
\caption{Quantitative comparison with state-of-the-art methods on the synthetic dataset.}
\label{tab:compare_sota_ourblender}
\end{table*}

\begin{table}[ht]
\centering
\caption{User study results (mean $\pm$ standard deviation).}
\label{tab:user_study}
\resizebox{0.45\textwidth}{!}{
    \begin{tabular}{lccc}
        \toprule
        Method & Vis. Qual. & Control & Consist. \\
        \midrule
        DaS & 1.82$\pm$1.06  &  1.48$\pm$0.76 & 1.57$\pm$0.90  \\
        TrajCrafter &  2.19 $\pm$1.11 & 2.20 $\pm$1.21 & 2.10 $\pm$ 1.17 \\
        ReCamMaster &  3.38$\pm$0.93 & 3.52$\pm$0.89 & 3.31$\pm$ 0.96  \\
        \textbf{PostCam} & \textbf{ 4.42$\pm$0.57} & \textbf{4.41$\pm$0.64} & \textbf{ 4.49$\pm$0.56} \\
        \bottomrule
    \end{tabular}
}
\end{table}

%% file: table/ablation_RT_Render_openvid.tex
\begin{table}[t]
\centering
\renewcommand{\arraystretch}{1.2}
\resizebox{0.5\textwidth}{!}{
\begin{tabular}{l|cc|cc}
\hline
\multirow{2}{*}{\textbf{Method}}
& \multicolumn{2}{c|}{\textbf{Camera Accuracy}}
& \multicolumn{2}{c}{\textbf{Video Quality}}\\
\cline{2-3}\cline{4-5}
    & \makecell[c]{RotErr$\downarrow$}
    & \makecell[c]{TransErr$\downarrow$}
    & \makecell[c]{Aes. Qual.$\uparrow$}
    & \makecell[c]{Img. Qual.$\uparrow$}\\
\hline
Pose-only
&
0.0841&
0.1545&
58.15&
63.63\\
Render-only
&
0.0548&
0.1786&
57.22&
62.63\\
Baseline Fusion(Pl\"ucker+Render)&
0.0727&
0.2312&
58.61&
\textbf{64.95}\\
Baseline Fusion(RT+Render)&
0.1140&
0.3718&
58.12&
62.73\\
PostCam(Pl\"ucker+Render)&
\textbf{0.0465} &
0.1259 &
58.70  &
64.83\\
PostCam(RT+Render)&
0.0495&
\textbf{0.0791}&
\textbf{59.80}&
64.64\\
\hline
\end{tabular}}
\caption{Ablation study on the synthetic dataset evaluating different camera conditioning strategies.
}
\label{tab:ablation_render_rt}
\end{table}

%% file: table/ablation_div_qkv_openvid.tex
\begin{table}[t]
\centering
\renewcommand{\arraystretch}{1.2}
\resizebox{0.45\textwidth}{!}{
\begin{tabular}{l|cc|cc}
\hline
\multirow{2}{*}{\textbf{Method}}
& \multicolumn{2}{c|}{\textbf{Camera Accuracy}}
& \multicolumn{2}{c}{\textbf{Video Quality}}\\
\cline{2-3}\cline{4-5}
    & \makecell[c]{RotErr$\downarrow$}
    & \makecell[c]{TransErr$\downarrow$}
    & \makecell[c]{Aes. Qual.$\uparrow$}
    & \makecell[c]{Img. Qual.$\uparrow$}\\
\hline
w/o query-shared&
0.1386&
0.1707&
59.13&
\textbf{65.94}\\
w/o kv-concatenated&
0.1389&
0.1984&
59.33&
65.90\\
Ours&
\textbf{0.0495}&
\textbf{0.0791}&
\textbf{59.80}&
64.64\\
\hline
\end{tabular}}
\caption{Ablation study on the synthetic dataset evaluating different attention injection mechanisms.
}
\label{tab:ablation_qs}
\end{table}

%% file: table/ablation_train_strategy.tex
\begin{table}[t]
\centering
\renewcommand{\arraystretch}{1.2}
\resizebox{0.45\textwidth}{!}{
\begin{tabular}{l|cc|cc}
\hline
\multirow{2}{*}{\textbf{Method}}
& \multicolumn{2}{c|}{\textbf{Camera Accuracy}}
& \multicolumn{2}{c}{\textbf{Video Quality}}\\
\cline{2-3}\cline{4-5}
    & \makecell[c]{RotErr$\downarrow$}
    & \makecell[c]{TransErr$\downarrow$}
    & \makecell[c]{Aes. Qual.$\uparrow$}
    & \makecell[c]{Img. Qual.$\uparrow$}\\
\hline
one stage&
0.0578&
0.1299&
58.98&
\textbf{65.05}\\
two stage, render-first&
0.0991&
0.1730&
56.89&
63.08\\
two stage, pose-first&
\textbf{0.0495}&
\textbf{0.0791}&
\textbf{59.80}&
64.64\\
\hline
\end{tabular}}
\caption{Ablation study on the synthetic dataset evaluating different training strategies.
}
\label{tab:ablation_train}
\end{table}

%% file: X_suppl.tex
\appendix

\section{Synthetic Dataset}
\label{sec:appendix_synthetic}
\begin{figure}[h!]
  \centering
  \includegraphics[width=0.65\linewidth]{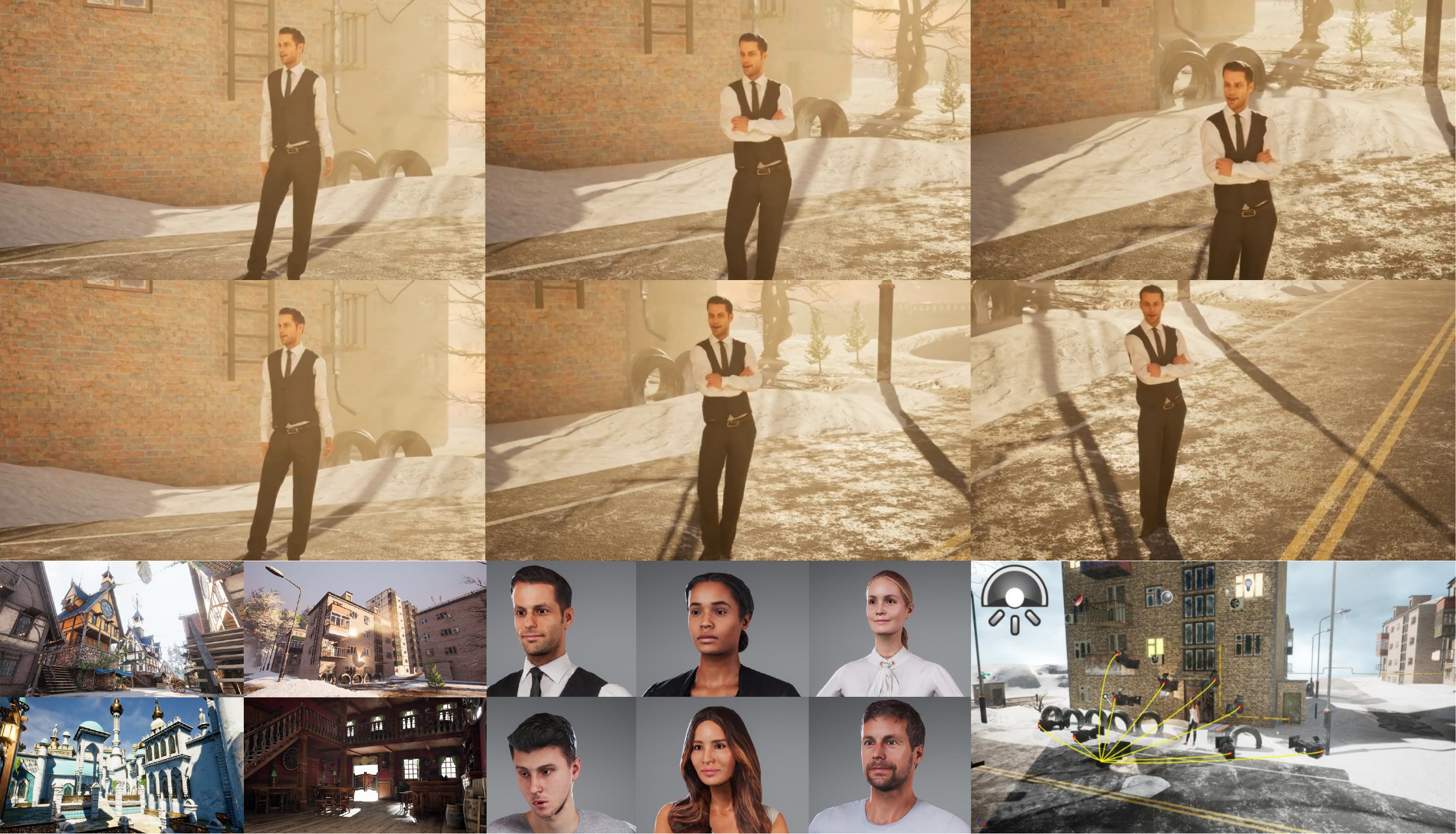}
  \caption{\textbf{Synthetic test dataset.}
The first two rows depict a dynamic scene from the synthetic test dataset under different camera trajectories. The final row presents representative examples of the scene layout, character appearance, and the intermediate rendering process.
  }
  \Description{Synthetic test dataset showing dynamic scenes under different camera trajectories, along with scene layout, character appearance, and rendering process examples.}
  \label{fig:blender}
\end{figure}

\begin{figure*}[ht!]
\centering
  \includegraphics[width=0.8\linewidth]{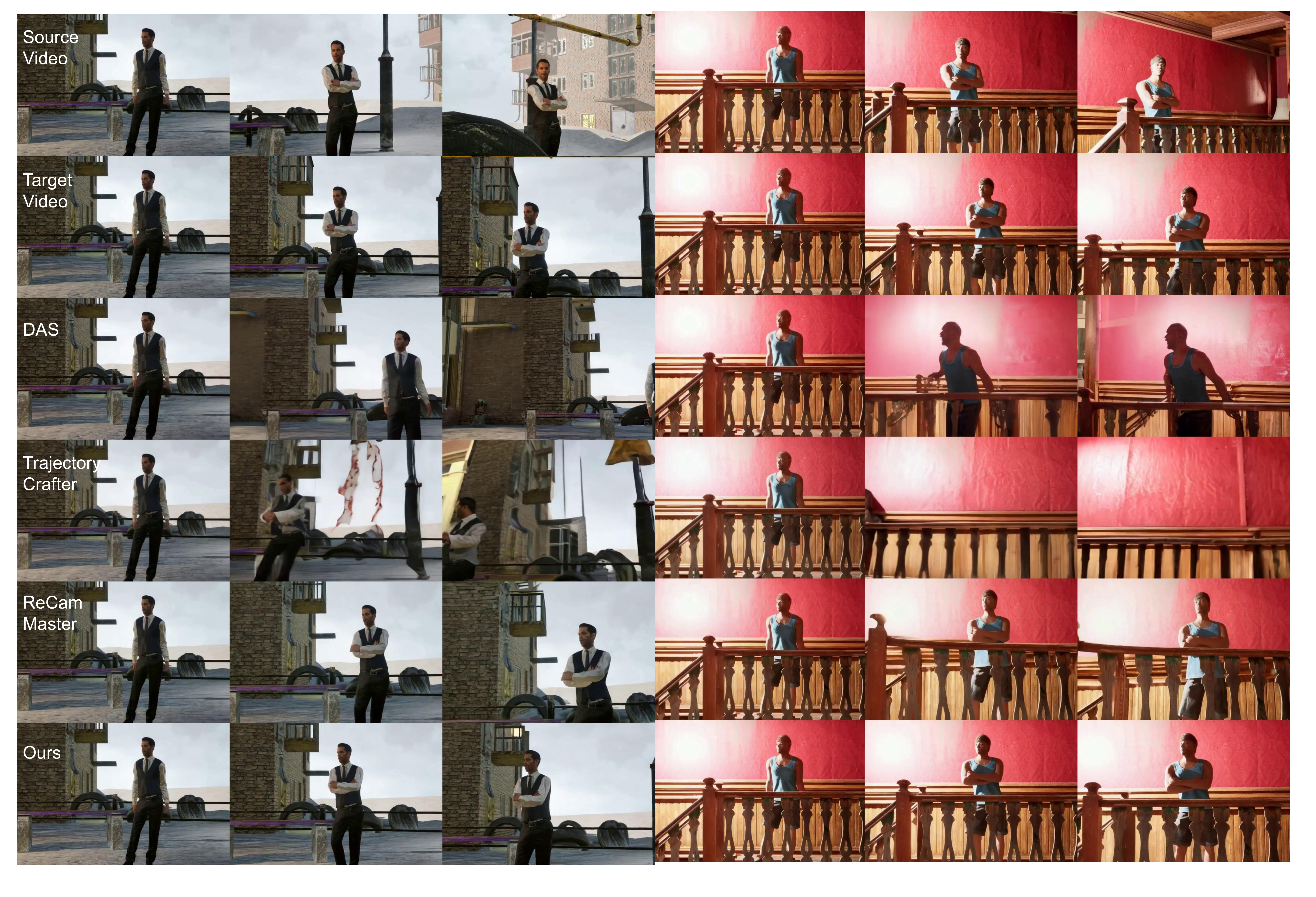}
  \caption{\textbf{Qualitative comparison results in synthetic dataset.}
Each row corresponds to a different method; the first two rows show the source video and the ground-truth target view, respectively. PostCam produces noticeably higher-quality results than other methods, and its alignment with the target view further highlights its superior camera control accuracy.
}
  \Description{Qualitative comparison on synthetic dataset showing source video, ground-truth target view, and results from different methods across multiple rows.}
  \label{fig:ue5}
\end{figure*}

Synchronized multi-view videos of dynamic scenes are difficult to capture in the real world.
To address this, we created a lightweight synthetic benchmark where each test sample (source video + target trajectory) pairs with a ground-truth target video rendered from the same scene and animation, differing only by camera viewpoint.
This design enables systematic evaluation of PostCam's camera-control accuracy. As can be seen in Fig.~\ref{fig:blender}, all characters and scenes originate from the official asset library of Unreal Engine 5, ultimately composing ten dynamic scenes featuring distinct styles and characters.
For each scene, we randomly sampled ten start–end camera poses within the navigable space and smoothly interpolated them with quadratic Bézier curves to produce 100-frame 1920×1080 trajectories; small Gaussian noise added to the control points further diversifies the motions.
Beyond RGB frames, we also rendered depth and normal maps for each frame. These were not used in our evaluation (where VGGT-estimated depth was used), but are provided for future research. The dataset and complete generation scripts will be publicly released.
Qualitative comparisons on the synthetic dataset are also presented in Fig.~\ref{fig:ue5}, where each row shows the source video, the ground-truth target view, and results from different methods. The results visually highlight PostCam's advantages in camera pose accuracy and cross-view consistency.

\section{Evaluation of Camera Pose Estimation Methods}
\label{sec:appendix_pose_eval}

In this section, we provide a detailed evaluation of various camera pose estimation methods to justify the selection of VGGT \cite{wang2025vggt} for our pipeline.

We conducted our evaluation on our synthetic test set. The protocol involved processing the videos with each pose estimation method to compute the camera trajectory. The predicted trajectory was then compared against the ground-truth camera trajectory provided by the dataset. We use standard metrics for evaluation: Rotational Error(denotes Rot. Err.) and Translational Error(denotes Trans. Err.).

We evaluated conventional
Structure-from-Motion methods, including COLMAP \cite{schoenberger2016sfm}, dynamic-scene-oriented models Pi3 \cite{wang2025pi3}, and VGGT \cite{wang2025vggt}.

We observed that conventional Structure-from-Motion methods, struggle significantly with the dynamic elements present in our test videos. These methods frequently failed to reconstruct the camera poses correctly, exhibiting a low success rate of approximately 30\%.

In contrast, both VGGT and Pi3 performed more robustly. As shown in Table~\ref{tab:pose_comparison}, reveals that VGGT is more suitable for our experiments. Based on this evidence, we selected VGGT as the core pose estimation method for all experiments presented in the main paper.

\begin{table}[h]
\centering
\caption{Comparison of Pose Estimation Accuracy on Synthetic Dynamic Scenes}
\label{tab:pose_comparison}
\begin{tabular}{l c c l}
\toprule
Method & Rot. Err. & Trans. Err. & Notes \\
\midrule
COLMAP & - & - & Low success rate ($\sim$30\%) \\
Pi3 & 0.0304 & 0.0352 & - \\
\textbf{VGGT (Ours)} & \textbf{0.0275} & \textbf{0.0241} & - \\
\bottomrule
\end{tabular}
\end{table}

\begin{figure*}[h!]
\centering
  \includegraphics[width=0.8\linewidth]{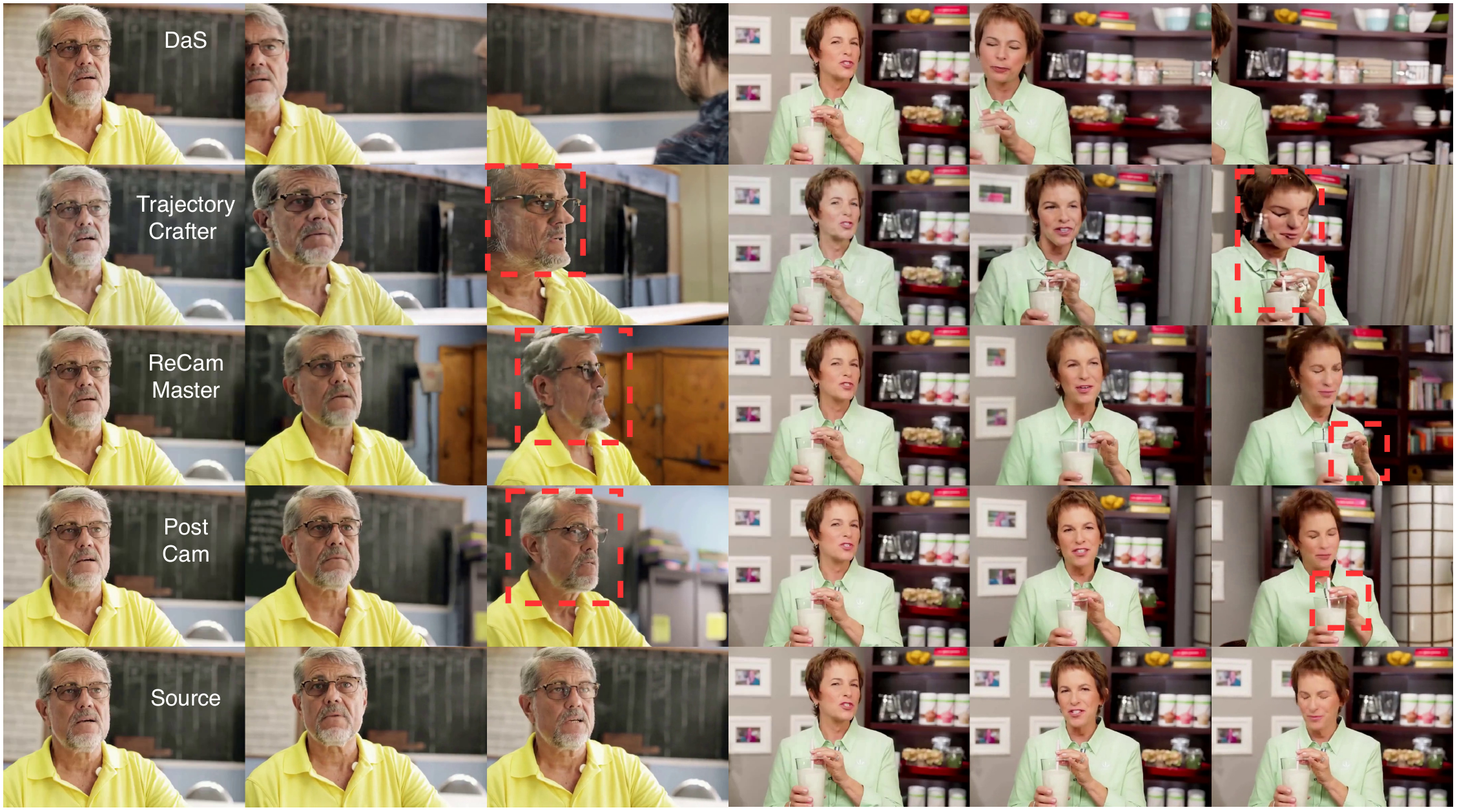}
  \caption{\textbf{Qualitative comparison results in OpenVid dataset.}
Two test sequences from the OpenVid dataset. Each row represents a different method, with the last row showing the original input video. In the first sequence, only our method preserves consistent facial features across frames, while other methods show identity shifts or visual artifacts. In the second sequence, only our method maintains detailed hand structures, avoiding the warping seen in baseline results. Red dashed boxes indicate regions that highlight the strengths of our approach.
}
  \Description{Qualitative comparison on OpenVid dataset showing two test sequences with results from different methods, highlighting facial feature preservation and hand structure consistency.}
  \label{fig:sota_compare}
\end{figure*}

\begin{figure*}[ht!]
\centering
  \includegraphics[width=0.9\linewidth]{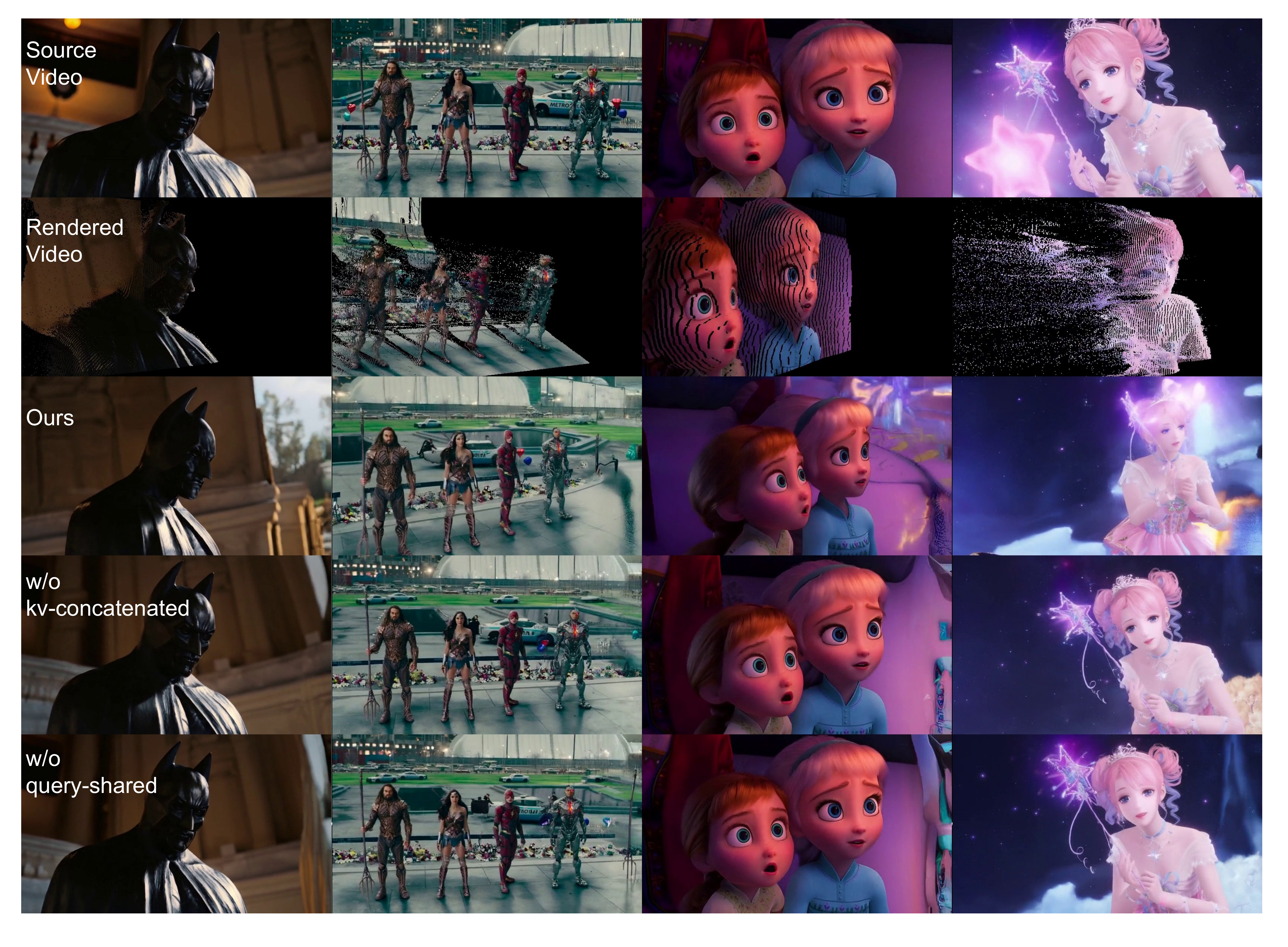}
  \caption{\textbf{Comparisons under different attention mechanism.}
Each row corresponds to a different method; the first two rows show the source videos and the rendered videos.
By examining the motion consistency with the rendered videos, it is evident that all ablated variants perform significantly worse than PostCam, showing the effectiveness of the proposed query-shared cross-attention in enhancing pose accuracy.
The higher image quality scores observed on VBench are primarily a consequence of poor camera control, which leads to reduced camera motion. With less camera movement, the generated video becomes more static, enabling a larger portion of the scene to be directly reused from the source video. This simplifies the generation task and artificially inflates the image quality metrics.
}
  \Description{Ablation study comparing different attention mechanisms, showing source videos, rendered videos, and results from ablated variants versus PostCam.}
  \label{fig:wo}
\end{figure*}



%% file: main.bib
@String(CVPR= {IEEE Conf. Comput. Vis. Pattern Recog.})

@String(CVPR  = {CVPR})

@article{blattmann2023stable,
  title={Stable video diffusion: Scaling latent video diffusion models to large datasets},
  author={Blattmann, Andreas and Dockhorn, Tim and Kulal, Sumith and Mendelevitch, Daniel and Kilian, Maciej and Lorenz, Dominik and Levi, Yam and English, Zion and Voleti, Vikram and Letts, Adam and others},
  journal={arXiv preprint arXiv:2311.15127},
  year={2023}
}

@article{guo2023animatediff,
  title={Animatediff: Animate your personalized text-to-image diffusion models without specific tuning},
  author={Guo, Yuwei and Yang, Ceyuan and Rao, Anyi and Liang, Zhengyang and Wang, Yaohui and Qiao, Yu and Agrawala, Maneesh and Lin, Dahua and Dai, Bo},
  journal={arXiv preprint arXiv:2307.04725},
  year={2023}
}

@article{singer2022make,
  title={Make-a-video: Text-to-video generation without text-video data},
  author={Singer, Uriel and Polyak, Adam and Hayes, Thomas and Yin, Xi and An, Jie and Zhang, Songyang and Hu, Qiyuan and Yang, Harry and Ashual, Oron and Gafni, Oran and others},
  journal={arXiv preprint arXiv:2209.14792},
  year={2022}
}

@article{ho2022video,
  title={Video diffusion models},
  author={Ho, Jonathan and Salimans, Tim and Gritsenko, Alexey and Chan, William and Norouzi, Mohammad and Fleet, David J},
  journal={Advances in Neural Information Processing Systems},
  volume={35},
  pages={8633--8646},
  year={2022}
}

@article{yang2024cogvideox,
  title={Cogvideox: Text-to-video diffusion models with an expert transformer},
  author={Yang, Zhuoyi and Teng, Jiayan and Zheng, Wendi and Ding, Ming and Huang, Shiyu and Xu, Jiazheng and Yang, Yuanming and Hong, Wenyi and Zhang, Xiaohan and Feng, Guanyu and others},
  journal={arXiv preprint arXiv:2408.06072},
  year={2024}
}

@article{brooks2024video,
  title={Video generation models as world simulators},
  author={Brooks, Tim and Peebles, Bill and Holmes, Connor and DePue, Will and Guo, Yufei and Jing, Li and Schnurr, David and Taylor, Joe and Luhman, Troy and Luhman, Eric and others},
  journal={OpenAI Blog},
  volume={1},
  pages={8},
  year={2024}
}

@article{wan2025wan,
  title={Wan: Open and advanced large-scale video generative models},
  author={Wan, Team and Wang, Ang and Ai, Baole and Wen, Bin and Mao, Chaojie and Xie, Chen-Wei and Chen, Di and Yu, Feiwu and Zhao, Haiming and Yang, Jianxiao and others},
  journal={arXiv preprint arXiv:2503.20314},
  year={2025}
}

@article{zheng2024open,
  title={Open-sora: Democratizing efficient video production for all},
  author={Zheng, Zangwei and Peng, Xiangyu and Yang, Tianji and Shen, Chenhui and Li, Shenggui and Liu, Hongxin and Zhou, Yukun and Li, Tianyi and You, Yang},
  journal={arXiv preprint arXiv:2412.20404},
  year={2024}
}

@article{kong2024hunyuanvideo,
  title={Hunyuanvideo: A systematic framework for large video generative models},
  author={Kong, Weijie and Tian, Qi and Zhang, Zijian and Min, Rox and Dai, Zuozhuo and Zhou, Jin and Xiong, Jiangfeng and Li, Xin and Wu, Bo and Zhang, Jianwei and others},
  journal={arXiv preprint arXiv:2412.03603},
  year={2024}
}

@inproceedings{wang2024motionctrl,
  title={Motionctrl: A unified and flexible motion controller for video generation},
  author={Wang, Zhouxia and Yuan, Ziyang and Wang, Xintao and Li, Yaowei and Chen, Tianshui and Xia, Menghan and Luo, Ping and Shan, Ying},
  booktitle={ACM SIGGRAPH 2024 Conference Papers},
  pages={1--11},
  year={2024}
}

@inproceedings{bahmani2025ac3d,
  title={Ac3d: Analyzing and improving 3d camera control in video diffusion transformers},
  author={Bahmani, Sherwin and Skorokhodov, Ivan and Qian, Guocheng and Siarohin, Aliaksandr and Menapace, Willi and Tagliasacchi, Andrea and Lindell, David B and Tulyakov, Sergey},
  booktitle={Proceedings of the Computer Vision and Pattern Recognition Conference},
  pages={22875--22889},
  year={2025}
}

@article{he2024cameractrl,
  title={Cameractrl: Enabling camera control for text-to-video generation},
  author={He, Hao and Xu, Yinghao and Guo, Yuwei and Wetzstein, Gordon and Dai, Bo and Li, Hongsheng and Yang, Ceyuan},
  journal={arXiv preprint arXiv:2404.02101},
  year={2024}
}

@article{bahmani2024vd3d,
  title={Vd3d: Taming large video diffusion transformers for 3d camera control},
  author={Bahmani, Sherwin and Skorokhodov, Ivan and Siarohin, Aliaksandr and Menapace, Willi and Qian, Guocheng and Vasilkovsky, Michael and Lee, Hsin-Ying and Wang, Chaoyang and Zou, Jiaxu and Tagliasacchi, Andrea and others},
  journal={arXiv preprint arXiv:2407.12781},
  year={2024}
}

@article{kuang2024collaborative,
  title={Collaborative video diffusion: Consistent multi-video generation with camera control},
  author={Kuang, Zhengfei and Cai, Shengqu and He, Hao and Xu, Yinghao and Li, Hongsheng and Guibas, Leonidas J and Wetzstein, Gordon},
  journal={Advances in Neural Information Processing Systems},
  volume={37},
  pages={16240--16271},
  year={2024}
}

@article{hou2024training,
  title={Training-free camera control for video generation},
  author={Hou, Chen and Wei, Guoqiang and Zeng, Yan and Chen, Zhibo},
  journal={arXiv preprint arXiv:2406.10126},
  year={2024}
}

@article{hu2024motionmaster,
  title={Motionmaster: Training-free camera motion transfer for video generation},
  author={Hu, Teng and Zhang, Jiangning and Yi, Ran and Wang, Yating and Huang, Hongrui and Weng, Jieyu and Wang, Yabiao and Ma, Lizhuang},
  journal={arXiv preprint arXiv:2404.15789},
  year={2024}
}

@article{ling2024motionclone,
  title={Motionclone: Training-free motion cloning for controllable video generation},
  author={Ling, Pengyang and Bu, Jiazi and Zhang, Pan and Dong, Xiaoyi and Zang, Yuhang and Wu, Tong and Chen, Huaian and Wang, Jiaqi and Jin, Yi},
  journal={arXiv preprint arXiv:2406.05338},
  year={2024}
}

@article{xiao2024video,
  title={Video diffusion models are training-free motion interpreter and controller},
  author={Xiao, Zeqi and Zhou, Yifan and Yang, Shuai and Pan, Xingang},
  journal={arXiv preprint arXiv:2405.14864},
  year={2024}
}

@article{yesiltepe2025dynamic,
  title={Dynamic View Synthesis as an Inverse Problem},
  author={Yesiltepe, Hidir and Yanardag, Pinar},
  journal={arXiv preprint arXiv:2506.08004},
  year={2025}
}

@article{zheng2024cami2v,
  title={Cami2v: Camera-controlled image-to-video diffusion model},
  author={Zheng, Guangcong and Li, Teng and Jiang, Rui and Lu, Yehao and Wu, Tao and Li, Xi},
  journal={arXiv preprint arXiv:2410.15957},
  year={2024}
}

@inproceedings{liang2025wonderland,
  title={Wonderland: Navigating 3d scenes from a single image},
  author={Liang, Hanwen and Cao, Junli and Goel, Vidit and Qian, Guocheng and Korolev, Sergei and Terzopoulos, Demetri and Plataniotis, Konstantinos N and Tulyakov, Sergey and Ren, Jian},
  booktitle={Proceedings of the Computer Vision and Pattern Recognition Conference},
  pages={798--810},
  year={2025}
}

@article{xu2024camco,
  title={Camco: Camera-controllable 3d-consistent image-to-video generation},
  author={Xu, Dejia and Nie, Weili and Liu, Chao and Liu, Sifei and Kautz, Jan and Wang, Zhangyang and Vahdat, Arash},
  journal={arXiv preprint arXiv:2406.02509},
  year={2024}
}

@article{yu2024viewcrafter,
  title={Viewcrafter: Taming video diffusion models for high-fidelity novel view synthesis},
  author={Yu, Wangbo and Xing, Jinbo and Yuan, Li and Hu, Wenbo and Li, Xiaoyu and Huang, Zhipeng and Gao, Xiangjun and Wong, Tien-Tsin and Shan, Ying and Tian, Yonghong},
  journal={arXiv preprint arXiv:2409.02048},
  year={2024}
}

@inproceedings{muller2024multidiff,
  title={Multidiff: Consistent novel view synthesis from a single image},
  author={M{\"u}ller, Norman and Schwarz, Katja and R{\"o}ssle, Barbara and Porzi, Lorenzo and Bul{\`o}, Samuel Rota and Nie{\ss}ner, Matthias and Kontschieder, Peter},
  booktitle={Proceedings of the IEEE/CVF Conference on Computer Vision and Pattern Recognition},
  pages={10258--10268},
  year={2024}
}

@inproceedings{ren2025gen3c,
  title={Gen3c: 3d-informed world-consistent video generation with precise camera control},
  author={Ren, Xuanchi and Shen, Tianchang and Huang, Jiahui and Ling, Huan and Lu, Yifan and Nimier-David, Merlin and M{\"u}ller, Thomas and Keller, Alexander and Fidler, Sanja and Gao, Jun},
  booktitle={Proceedings of the Computer Vision and Pattern Recognition Conference},
  pages={6121--6132},
  year={2025}
}

@article{li2025realcam,
  title={Realcam-i2v: Real-world image-to-video generation with interactive complex camera control},
  author={Li, Teng and Zheng, Guangcong and Jiang, Rui and Wu, Tao and Lu, Yehao and Lin, Yining and Li, Xi and others},
  journal={arXiv preprint arXiv:2502.10059},
  year={2025}
}

@article{feng2024i2vcontrol,
  title={I2VControl-Camera: Precise Video Camera Control with Adjustable Motion Strength},
  author={Feng, Wanquan and Liu, Jiawei and Tu, Pengqi and Qi, Tianhao and Sun, Mingzhen and Ma, Tianxiang and Zhao, Songtao and Zhou, Siyu and He, Qian},
  journal={arXiv preprint arXiv:2411.06525},
  year={2024}
}

@article{popov2025camctrl3d,
  title={CamCtrl3D: Single-Image Scene Exploration with Precise 3D Camera Control},
  author={Popov, Stefan and Raj, Amit and Krainin, Michael and Li, Yuanzhen and Freeman, William T and Rubinstein, Michael},
  journal={arXiv preprint arXiv:2501.06006},
  year={2025}
}

@article{zhai2025stargen,
  title={StarGen: A Spatiotemporal Autoregression Framework with Video Diffusion Model for Scalable and Controllable Scene Generation},
  author={Zhai, Shangjin and Ye, Zhichao and Liu, Jialin and Xie, Weijian and Hu, Jiaqi and Peng, Zhen and Xue, Hua and Chen, Danpeng and Wang, Xiaomeng and Yang, Lei and others},
  journal={arXiv preprint arXiv:2501.05763},
  year={2025}
}

@inproceedings{van2024generative,
  title={Generative camera dolly: Extreme monocular dynamic novel view synthesis},
  author={Van Hoorick, Basile and Wu, Rundi and Ozguroglu, Ege and Sargent, Kyle and Liu, Ruoshi and Tokmakov, Pavel and Dave, Achal and Zheng, Changxi and Vondrick, Carl},
  booktitle={European Conference on Computer Vision},
  pages={313--331},
  year={2024},
  organization={Springer}
}

@article{bai2025recammaster,
  title={Recammaster: Camera-controlled generative rendering from a single video},
  author={Bai, Jianhong and Xia, Menghan and Fu, Xiao and Wang, Xintao and Mu, Lianrui and Cao, Jinwen and Liu, Zuozhu and Hu, Haoji and Bai, Xiang and Wan, Pengfei and others},
  journal={arXiv preprint arXiv:2503.11647},
  year={2025}
}

@inproceedings{zhang2025recapture,
  title={Recapture: Generative video camera controls for user-provided videos using masked video fine-tuning},
  author={Zhang, David Junhao and Paiss, Roni and Zada, Shiran and Karnad, Nikhil and Jacobs, David E and Pritch, Yael and Mosseri, Inbar and Shou, Mike Zheng and Wadhwa, Neal and Ruiz, Nataniel},
  booktitle={Proceedings of the Computer Vision and Pattern Recognition Conference},
  pages={2050--2062},
  year={2025}
}

@article{gu2025diffusion,
  title={Diffusion as Shader: 3D-aware Video Diffusion for Versatile Video Generation Control},
  author={Gu, Zekai and Yan, Rui and Lu, Jiahao and Li, Peng and Dou, Zhiyang and Si, Chenyang and Dong, Zhen and Liu, Qifeng and Lin, Cheng and Liu, Ziwei and others},
  journal={arXiv preprint arXiv:2501.03847},
  year={2025}
}

@article{xiao2024trajectory,
  title={Trajectory Attention for Fine-grained Video Motion Control},
  author={Xiao, Zeqi and Ouyang, Wenqi and Zhou, Yifan and Yang, Shuai and Yang, Lei and Si, Jianlou and Pan, Xingang},
  journal={arXiv preprint arXiv:2411.19324},
  year={2024}
}

@article{bian2025gs,
  title={GS-DiT: Advancing Video Generation with Pseudo 4D Gaussian Fields through Efficient Dense 3D Point Tracking},
  author={Bian, Weikang and Huang, Zhaoyang and Shi, Xiaoyu and Li, Yijin and Wang, Fu-Yun and Li, Hongsheng},
  journal={arXiv preprint arXiv:2501.02690},
  year={2025}
}

@article{mark2025trajectorycrafter,
  title={Trajectorycrafter: Redirecting camera trajectory for monocular videos via diffusion models},
  author={Mark, YU and Hu, Wenbo and Xing, Jinbo and Shan, Ying},
  journal={arXiv preprint arXiv:2503.05638},
  volume={2},
  year={2025}
}

@article{lipman2022flow,
  title={Flow matching for generative modeling},
  author={Lipman, Yaron and Chen, Ricky TQ and Ben-Hamu, Heli and Nickel, Maximilian and Le, Matt},
  journal={arXiv preprint arXiv:2210.02747},
  year={2022}
}

@inproceedings{wang2025vggt,
  title={Vggt: Visual geometry grounded transformer},
  author={Wang, Jianyuan and Chen, Minghao and Karaev, Nikita and Vedaldi, Andrea and Rupprecht, Christian and Novotny, David},
  booktitle={Proceedings of the Computer Vision and Pattern Recognition Conference},
  pages={5294--5306},
  year={2025}
}

@article{nan2024openvid,
  title={Openvid-1m: A large-scale high-quality dataset for text-to-video generation},
  author={Nan, Kepan and Xie, Rui and Zhou, Penghao and Fan, Tiehan and Yang, Zhenheng and Chen, Zhijie and Li, Xiang and Yang, Jian and Tai, Ying},
  journal={arXiv preprint arXiv:2407.02371},
  year={2024}
}

@inproceedings{huang2024vbench,
  title={Vbench: Comprehensive benchmark suite for video generative models},
  author={Huang, Ziqi and He, Yinan and Yu, Jiashuo and Zhang, Fan and Si, Chenyang and Jiang, Yuming and Zhang, Yuanhan and Wu, Tianxing and Jin, Qingyang and Chanpaisit, Nattapol and others},
  booktitle={Proceedings of the IEEE/CVF Conference on Computer Vision and Pattern Recognition},
  pages={21807--21818},
  year={2024}
}

@article{heusel2017gans,
  title={Gans trained by a two time-scale update rule converge to a local nash equilibrium},
  author={Heusel, Martin and Ramsauer, Hubert and Unterthiner, Thomas and Nessler, Bernhard and Hochreiter, Sepp},
  journal={Advances in neural information processing systems},
  volume={30},
  year={2017}
}

@inproceedings{peebles2023scalable,
  title={Scalable diffusion models with transformers},
  author={Peebles, William and Xie, Saining},
  booktitle={Proceedings of the IEEE/CVF international conference on computer vision},
  pages={4195--4205},
  year={2023}
}

@inproceedings{cao2025mvgenmaster,
  title={MVGenMaster: Scaling Multi-View Generation from Any Image via 3D Priors Enhanced Diffusion Model},
  author={Cao, Chenjie and Yu, Chaohui and Liu, Shang and Wang, Fan and Xue, Xiangyang and Fu, Yanwei},
  booktitle={Proceedings of the Computer Vision and Pattern Recognition Conference},
  pages={6045--6056},
  year={2025}
}

@article{cao2025uni3c,
  title={Uni3C: Unifying Precisely 3D-Enhanced Camera and Human Motion Controls for Video Generation},
  author={Cao, Chenjie and Zhou, Jingkai and Li, Shikai and Liang, Jingyun and Yu, Chaohui and Wang, Fan and Xue, Xiangyang and Fu, Yanwei},
  journal={arXiv preprint arXiv:2504.14899},
  year={2025}
}

@inproceedings{schoenberger2016sfm,
    author={Sch\"{o}nberger, Johannes Lutz and Frahm, Jan-Michael},
    title={Structure-from-Motion Revisited},
    booktitle={Conference on Computer Vision and Pattern Recognition (CVPR)},
    year={2016},
}

@misc{wang2025pi3,
      title={$\pi^3$: Scalable Permutation-Equivariant Visual Geometry Learning}, 
      author={Yifan Wang and Jianjun Zhou and Haoyi Zhu and Wenzheng Chang and Yang Zhou and Zizun Li and Junyi Chen and Jiangmiao Pang and Chunhua Shen and Tong He},
      year={2025},
      eprint={2507.13347},
      archivePrefix={arXiv},
      primaryClass={cs.CV},
      url={https://arxiv.org/abs/2507.13347}, 
}

@article{wang2024cpa,
  title={CPA: Camera-pose-awareness diffusion transformer for video generation},
  author={Wang, Yuelei and Zhang, Jian and Jiang, Pengtao and Zhang, Hao and Chen, Jinwei and Li, Bo},
  journal={arXiv preprint arXiv:2412.01429},
  year={2024}
}

@article{he2025cameractrl,
  title={Cameractrl ii: Dynamic scene exploration via camera-controlled video diffusion models},
  author={He, Hao and Yang, Ceyuan and Lin, Shanchuan and Xu, Yinghao and Wei, Meng and Gui, Liangke and Zhao, Qi and Wetzstein, Gordon and Jiang, Lu and Li, Hongsheng},
  journal={arXiv preprint arXiv:2503.10592},
  year={2025}
}

@article{unterthiner2018fvd,
  title={FVD: A New Metric for Video Generation},
  author={Unterthiner, Thomas and van Steenkiste, Sjoerd and Kurach, Karol and Marinier, Raphael and Michalski, Marcin and Gelly, Sylvain},
  journal={arXiv preprint arXiv:1812.01717},
  year={2018}
}
